\documentclass[times, review, 10pt]{elsarticle}

\usepackage{amssymb}
\usepackage{amsmath}
\usepackage{amsmath,amsfonts}
\usepackage{dsfont}
\newcommand{\mathbbm}[1]{\mathds{#1}}
\usepackage[caption=false,font=footnotesize,labelfont=sf,textfont=sf]{subfig}
\def\dif{\mathop{}\hphantom{\mskip-\thinmuskip}\mathrm{d}}
\usepackage{balance}
\usepackage[table,xcdraw]{xcolor}
\usepackage{booktabs}
\usepackage{hyperref}
\usepackage{multirow}
\usepackage{bbding}
\newcommand{\tabincell}[2]{\begin{tabular}{@{}#1@{}}#2\end{tabular}}
\journal{arXiv}

\begin{document}

\begin{frontmatter}



\title{WCCNet: Wavelet-context Cooperative Network for Efficient Multispectral Pedestrian Detection}


\author[cse]{Xingjian~Wang} %% Author name
\ead{xingjianwang@zju.edu.cn}
\author[cse]{Li~Chai\corref{cor1}} %% Author name
\ead{chaili@zju.edu.cn}
\author[cse]{Jiming~Chen} %% Author name
\ead{cjm@zju.edu.cn}
\author[isee]{Zhiguo~Shi} %% Author name
\ead{shizg@zju.edu.cn}

\affiliation[cse]{organization={the College of Control Science and Engineering, Zhejiang University},
            addressline={38 Zheda Road},
            city={Hangzhou},
            postcode={310027},
            state={Zhejiang},
            country={China}}
\affiliation[isee]{organization={the College of Information Science and Electronic Engineering, Zhejiang University},
            addressline={38 Zheda Road}, 
            city={Hangzhou},
            postcode={310027}, 
            state={Zhejiang},
            country={China}}
\cortext[cor1]{Corresponding author}

%% Abstract
\begin{abstract}
Multispectral pedestrian detection is essential to various tasks especially autonomous driving, for which both the accuracy and computational cost are of paramount importance. 
Most existing approaches treat RGB and infrared modalities equally.
They typically adopt two symmetrical backbones for multimodal feature extraction,  
which ignore the substantial differences between modalities and bring great difficulty for the reduction of the computational cost as well as effective crossmodal fusion.
In this work, we propose a novel and efficient framework named Wavelet-context Cooperative Network (WCCNet), which differentially extracts complementary features across spectra with low computational cost and further fuses these diverse features based on their spatially relevant cross-modal semantics.
WCCNet explores an asymmetric but cooperative dual-stream backbone, in which WCCNet utilizes generic neural layers for texture-rich feature extraction from RGB modality, while proposing Mixture of Wavelet Experts (MoWE) to capture complementary frequency patterns of infrared modality.
By assessing multispectral environmental context, MoWE generates routing scores to selectively activate specific learnable Adaptive DWT (ADWT) layers, alongside shared static DWT, which are both considerible lightwight and efficient.
This cooperative structure not only significantly reduces the computational complexity, but also facilitates the subsequent crossmodal fusion.
To further fuse these multispectral features with significant semantic differences, we elaborately design the crossmodal rearranging fusion module (CMRF), which aims to mitigate misalignment and merge semantically complementary features in spatially-related local regions to amplify the crossmodal reciprocal information.
Results from comprehensive evaluations on KAIST and FLIR benchmarks indicate that WCCNet outperforms state-of-the-art methods with considerable computational efficiency and competitive accuracy.
\end{abstract}


% \begin{highlights}
% \item Asymmetric backbone of WCCNet efficiently extracts spectral specific characteristics.
% \item WCCNet explores semantic-aware reciprocal relationship between two spectra.
% \item Lightweight MoWE captures complementary infrared features according to global context.
% \item CMRF generates on-the-fly attention weights to integrate spatially relevant features.
% \item WCCNet outperforms prior works with considerable detection accuracy and efficiency.

% \end{highlights}

%% Keywords

\end{frontmatter}


\section{Introduction\label{sec: intro}}
Pedestrian detection is a crucial task widely studied for autonomous driving owing to its central role in ensuring safety and preventing accidents.
Due to the limitations of RGB cameras in capturing pedestrian features under low-light environments, the integration of infrared modality with RGB modality \cite{MPD-1,MPD-2} considerably enhances the reliability \cite{KAIST2015,FLIRDataset}, and attracts growing research interests.
Recent years have witnessed rapid development of deep learning-based multispectral pedestrian detection methods.

Detection speed and accuracy are equally critical for time-sensitive autonomous driving scenario, since delayed or low-precision detection will both lead to severe consequences.
Therefore, it is vital to explore lightweight and efficient frameworks that also provide competitive accuracy.
However, the substantial differences in feature representation and semantic information between modalities pose a great challenge to fast feature extraction and effective crossmodal fusion.

Existing multispectral detection methods 
\cite{MSDS2018,ARCNN2019,CIAN2019,MBNet2020,MLPD2021,CMPD2022,liu2016CleanKAIST,RPN2017Konig,tang2022piafusion,wang2022RISNet,zhang2023TripleNet,TCDET2020,yang2022BAANet,ICAFusion} equally treat the feature extraction of RGB and infrared modalities, and typically apply symmetrical backbones for both RGB and infrared modalities.
This symmetry ignores the inherent disparity between spectra, and will inevitably entail extra computational burden.
Therefore, it is quite significant and challenging to find a new network with fewer parameters and high accuracy simultaneously. 
On the other hand, infrared images and RGB images have clearly distinct features \cite{CMPD2022,yang2022BAANet}.
RGB images have richer information about textures and colors \cite{MD-1}, which necessitate dense network to resolve.
Conversely, infrared images lack internal texture, and mainly contain valuable geometry details including low-frequency thermal blobs and high-frequency contours.
Processing infrared stream with heavyweight neural layers results in significant computational redundancy, and may overfit to thermal sensor noise like quantization artifacts.
It is suggested to design dynamic frequency-domain filters for efficient infrared feature extraction.

\begin{figure}[t]
  \centering
  \includegraphics[width=0.8\textwidth]{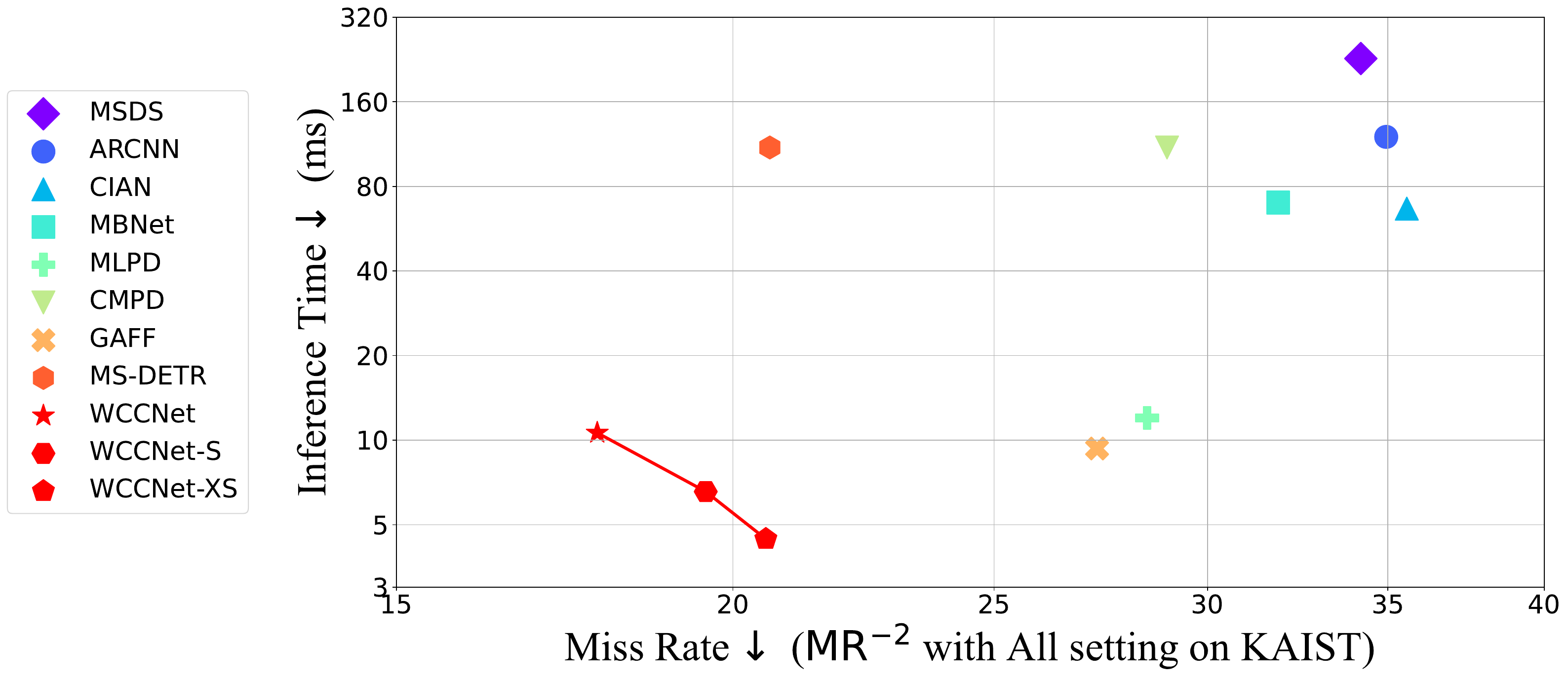}
  \caption{Inference time (ms) vs. $\mathrm{MR}^{-2}$ comparisons with state-of-the-art methods on the KAIST \cite{KAIST2015} test set. MSDS \cite{MSDS2018}, ARCNN \cite{ARCNN2019}, CIAN \cite{CIAN2019}, MBNet \cite{MBNet2020}, MLPD \cite{MLPD2021}, CMPD \cite{CMPD2022}, GAFF \cite{GAFF2021}, and MS-DETR \cite{MS-DETR2024} are involved. WCCNet, WCCNet-S, and WCCNet-XS are three variants with different amounts of parameters. Lower $\mathrm{MR}^{-2}$ represents higher precision, lower inference time denotes higher detection speed.
  \label{fig. Speed vs. Accuracy}
}
\end{figure}

Furthermore, the obvious visual differences between RGB and infrared modalities also signify a huge semantic distinction between their extracted features, which has posed a challenge to crossmodal fusion.
Conventional fusion schemes \cite{MSDS2018,ARCNN2019,MLPD2021,RPN2017Konig} directly apply simple pixel-wise addition and channel-dimension concatenation, thus they are weak in extracting complementary features.
Recent fusion schemes \cite{MBNet2020,tang2022piafusion,wang2022RISNet,zhang2023TripleNet} differentiate complementary features from redundant features along channel dimension based on the differences in their numeric values.
They assume explicitly or implicitly that features of different modalities are numerically homogeneous.
However, it is well-known that the semantic differences between features of different modalities cannot be adequately represented by their numeric differences, especially when there exists a huge semantic gap between them.
Since pedestrians in the paired RGB and infrared images are spatially related, we manage to explore semantic correlation between crossmodal features based on their inherent correlation.

To tackle with abovementioned limitations, we propose \textbf{W}avelet-\textbf{C}ontext \textbf{C}ooperative framework (WCCNet) for faster and more accurate multispectral pedestrian detection, consisting of MoWE-integrated cooperative backbone and crossmodal rearranging fusion module. 
WCCNet asymmetrically extracts features for infrared and RGB modalities with lower computational cost, and effectively aggregates semantically complementary features through crossmodal context in spatially relevant local regions.

\begin{figure}[t]
  \centering
  \includegraphics[width=\textwidth]{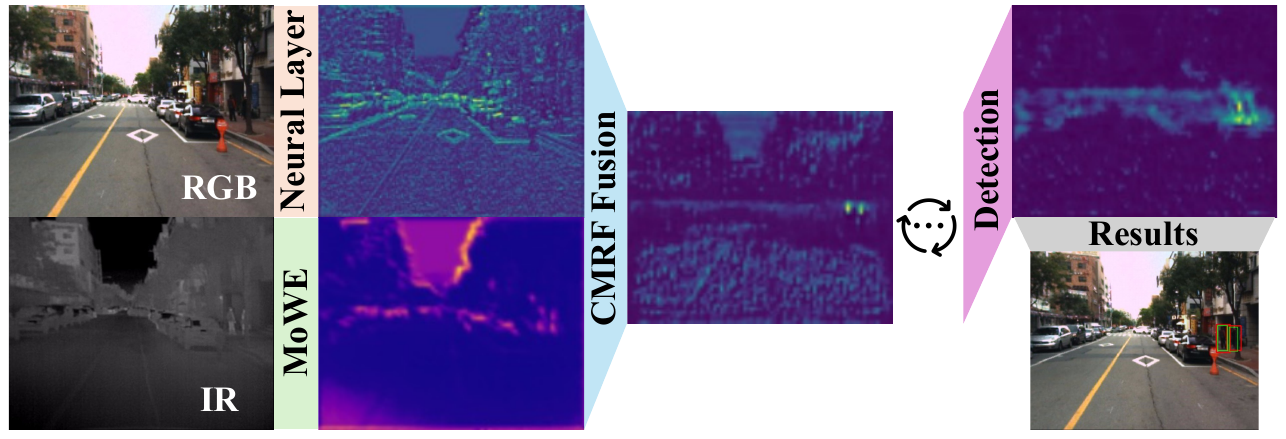}
  \caption{Feature visualization of WCCNet on KAIST testing set. For complex illumination scenarios, WCCNet effectively captures and integrates complementary features by wavelet-context cooperative framework.\label{fig. vis_feature}
}
\end{figure}

Firstly, we propose a wavelet-cooperative backbone that combines Mixture of Wavelet Experts (MoWE)–based infrared branch with generic neural layer-based RGB branch.
MoWE is designed to route channel-wise features to a diverse bank of wavelet filters, including a global context-aware router and two kinds of wavelet experts.
By assessing environmental context jointly from multispectral inputs, MoWE first produces routing scores that selectively activate a sparse subset of routed experts.
We propose Adaptive Discrete Wavelet Transform (ADWT) layers as routed experts, alongside with static DWT layers as shared experts that are always activated.
ADWT experts are constructed as learnable parallel DWT layers with adaptive weighted scores, allowing optimization of wavelet basis to better capture pedestrian-specific frequency patterns.
Complementing these are shared DWT experts utilizing pre-defined static weights as stable frequency prior.
MoWE is lightweight compared to generic neural layers, and its sparse activation further enhances inference efficiency.
As for RGB branch, we pratically employ convolutional neural network (CNN) as generic neural layers.
WCCNet framework also supports transformer-based layers as generic neural layers, e.g., ViT and Swin Transformer, but have not been explored in this paper.
The reciprocal relationship between MoWE and generic neural layers not only significantly reduces computational cost, but also yields informative infrared features to complement RGB modality and facilitate the subsequent crossmodal fusion, as shown in Fig. \ref{fig. vis_feature}.

Secondly, we propose CrossModal Rearranging Fusion (CMRF) module to fully leverage the semantic interaction of crossmodal features based on their spatial correlation.
CMRF first narrows the semantic gap between multimodal features extracted in different domains by the Cross-domain Embedding layer (CE), and spatially align them by the Crossmodal Spatial Alignment layer (CSA) to lessen the spatial misalignments of multimodal features.
Based on the mapping relationship between corresponding local regions of the two modalities, CMRF generates on-the-fly attention weights that focus on the locations of pedestrians.
These semantic-aware attention weights are utilized to transfer local instance-specific features as complementary information.
The proposed CMRF improves the accuracy and robustness of pedestrian detection.

The framework of WCCNet is illustrated in Fig. \ref{Fig.framework} and Fig. \ref{Fig. CMRF}, and our main contributions are summarized as follows:
\begin{itemize}
  \item A novel framework WCCNet is proposed to achieve considerable detection efficiency and competitive accuracy.
  WCCNet employs an asymmetric wavelet-cooperative backbone, utilizing lightweight MoWE to dynamically routes infrared inputs through activated ADWT experts to capture informative frequency-domain features, while extracting RGB features by generic neural layers.
  It leverages reciprocal relationship between MoWE and generic neural layers to extract complementary features with lower computational cost.

  \item CMRF is proposed to fuse semantically complementary and spatially relevant features of the two modalities with large discrepancy, generating on-the-fly attention weights to focus on spatially related and semantically complementary regions.
  Cross-domain Embedding (CE) and Crossmodal Spatial Alignment (CSA) layers are integrated into CMRF module to tackle the cross-domain semantic gap and the crossmodal spatial misalignment respectively.

  \item We conducted extensive experiments on KAIST and FLIR, two widely-used benchmarks on autonomous driving scenes.
  Experimental results demonstrate the superiority of WCCNet in detection efficiency, achieving high precision with significant reduction in learnable parameters and computational cost.
\end{itemize}

\section{Related Work}
\subsection{Multispectral Pedestrian Detection}
Multispectral pedestrian detection has attracted a growing research interest for its essential role in autonomous driving.
In the past few years, deep learning-based methods \cite{MSDS2018,ARCNN2019,CIAN2019,MBNet2020,MLPD2021,CMPD2022,liu2016CleanKAIST,RPN2017Konig,tang2022piafusion,wang2022RISNet,zhang2023TripleNet,TCDET2020,yang2022BAANet} have exhibited state-of-the-art performance and become the mainstream.
Most of them typically utilize a dual-stream framework to extract RGB and infrared features both by symmetrical neural layers.
% , while focussing on the multimodal feature extraction and crossmodal feature fusion.
Liu \emph{et al}. \cite{liu2016CleanKAIST} employed four fusion architectures which fuse at different stages of feature extraction, and found that the middle fusion produced the highest accuracy.
Based on this observation, MSDS \cite{MSDS2018} further improved the detection accuracy by refining the detection results using cascaded classification subnetwork.
% Considering position shift problem in weakly aligned multispectral images, AR-CNN \cite{ARCNN2019} designed an aligned region network to adaptively align the multimodal features. 
Considering totally unpaired multispectral images, MLPD\cite{MLPD2021} leveraged multi-label learning to obtain competitive detection accuracy in unpaired conditions.
As for crossmodal feature fusion, beyond conventional fusion methods including pixel-wise addition \cite{ARCNN2019} and channel-dimension concatenation \cite{MSDS2018,MLPD2021,RPN2017Konig}, 
MBNet \cite{MBNet2020} presented a differential modality aware fusion (DMAF) module for better complementary feature extraction.
Following the DMAF paradigm of MBNet, there witnessed a series of similar methods that achieved state-of-the-art performance, \emph{e.g.}, PIAFusion \cite{tang2022piafusion}, RISNet \cite{wang2022RISNet}, and TINet \cite{zhang2023TripleNet}, etc.
Employing an iterative cross-attention mechanism, ICAFusion \cite{ICAFusion} progressively refine and fuse complementary multispectral features.
Besides, MS-DETR \cite{MS-DETR2024} explored loosely coupled fusion strategy within a multi-modal transformer decoder.
In addition to feature-level fusion, CMPD \cite{CMPD2022} conducted decision-level fusion based on Dempster's combination rule.
% YOLO-MSLite\cite{YOLO-MSLite} further explores to improve detection efficiency.
% TFNet \cite{TFNet2024} further explores cross-dataset generalization ability based on thermal-first framework.

However, these prior works have neglected the intrinsic feature differences between RGB and infrared modalities, and simply utilized symmetrical backbones for both modalities.
In fact, for infrared modality rich in contour features but deficient in texture features, it is unnecessary to apply heavyweight neural layers at shallow stages, which may results in excessive computational burden and even feature redundancy.
Instead, we explore to extract features differentially according to the characteristics of different modalities with lower computational burden.

Meanwhile, crossmodal fusion schemes in prior works \cite{MBNet2020,tang2022piafusion,wang2022RISNet,zhang2023TripleNet} assumed features of different spectra are numerically homogeneous. 
Accordingly, they tended to filter out redundant features with minor numeric differences from multispectral features.
However, due to the huge semantic gap between modalities, redundant features of different modalities can present a large discrepancy in their numeric values, which will hinder the crossmodal fusion of these methods.
Therefore, in this study, we establish the semantic-aware interaction between crossmodal features based on their spatial consistency, and further exploit their complementary information.

\subsection{Wavelets in Deep Learning}
As an effective tool in time-frequency analysis, wavelets have been widely applied in various signal processing applications \cite{mallat1989wavelet,mallatWaveletBook}. 
A great deal of research has been devoted to the combination of wavelet and deep learning networks for both low-level and high-level visual tasks, demonstrating that superior results can be achieved by such methods. 
Early works primarily explored the integration of fixed wavelet transforms into deep networks. 
It is natural to design networks learning directly from the frequency domain \cite{deepinwavelet}. 
Further, in this context, wavelet scattering network utilizes fixed wavelet transform convolutions to cascade with nonlinear activations and averaging operators \cite{MallatDWT}. 
Fujieda et al. \cite{WaveletConvNet} proposed a structure to introduce multiscale frequency components at different stages, while Li et al. \cite{li2021wavecnet} proposed WaveCNet to suppress aliasing effects by replacing down-sampling operations with fixed DWTs, achieving better noise-robustness. 
More recently, research attention has shifted towards fully learnable wavelets. 
Michau et al. \cite{GabrielPNAS} construct learnable wavelet following conjugate quadrature filter bank property, showing effectiveness for unsupervised monitoring of high-frequency time series. 
Furthermore, the adaptive wavelet distillation (AWD) method \cite{WaveletDistill} learns an effective wavelet transform by penalizing feature attributions in the wavelet domain, yielding SOTA performance for cosmology inferring and molecular-partner prediction.
Motivated by inherent high efficiency of wavelet transform, we further explore a dynamic mixture mechanism of fixed and learnable wavelets to facilitate the extraction of diverse frequency characteristics for infrared modality while maintaining computational economy.

\subsection{Multispectral Fusion Scheme}
Fusion between feature maps of different semantic contents is essential for multispectral detection task.
Conventional fusion schemes regard input features as a whole and conduct simple numeric combinations, \emph{e.g.}, pixel-wise addition, pixel-wise multiplication, and channel-dimension concatenation, which are incapable of highlighting the crucial parts of input features.
In recent years, fusion schemes based on attention mechanisms have emerged.
SENet \cite{hu2018squeeze} applied channel attention with channel-wise weights from spatial global information, while CBAM \cite{woo2018cbam} additionally adopted spatial attention with a score map calculated from pixel-wise channel pooling.
The variants of the aforementioned channel and spatial attention mechanisms were widely used as the fusion schemes of prior multispectral detection works \cite{MBNet2020,tang2022piafusion,wang2022RISNet,zhang2023TripleNet,TCDET2020,yang2022BAANet}.

Recently, there are some works about learning dynamic kernels which may be useful for fusion tasks.
CARAFE++ \cite{wang2021carafe++} and IndexNet \cite{lu2022index} were proposed to generate adaptive kernels to aggregate contextual information within related receptive filed.
However, for multispectral features with a large semantic gap, these single-modal fusion schemes need pre-concatenation for multimodal input features, and thus can not capture contextual correlations between feature maps of different modalities.
Besides, these fusion schemes heavily rely on pixel-wise or channel-wise alignment of input features, thus prone to feature misalignment which are common in practical applications.
Instead, the contextual correlations between multimodal features of different modalities and the adaptive spatial feature alignment are explored in this work.

\section{Proposed Method}

% The overall architecture of WCCNet is shown in Fig. \ref{Fig.framework}. 
% In the following subsections, we shall mainly introduce the proposed backbone and fusion mechanism. 

\subsection{MoWE-integrated Wavelet-cooperative Backbone\label{Sec.backbone}}
\subsubsection{Overall Structure} 
\begin{figure}[!t]
  \centering
  \includegraphics[width=\textwidth]{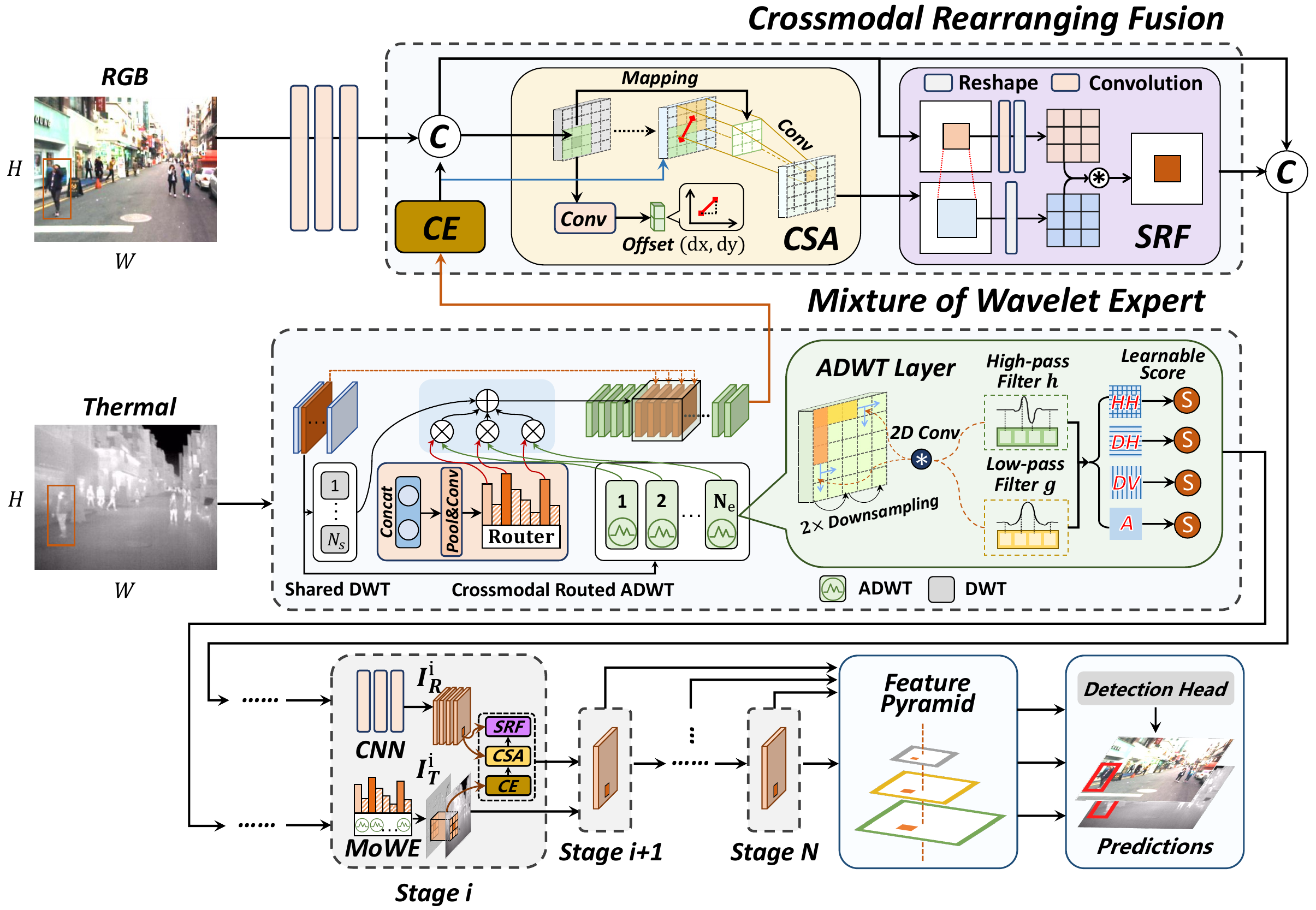}
  \caption{Overview framework of WCCNet. WCCNet mainly consists of three parts: (1) dual-stream backbone with MoWE and generic neural layers for fast and lightweight multispectral feature extraction; (2) CMRF module for semantic-aware crossmodal fusion; (3) decoupling detection head.\label{Fig.framework}}
\end{figure}

For higher computational efficiency and precision, we propose an asymmetric wavelet-cooperative backbone that extracts features differentially according to the characteristics of different modalities, as shown in Fig. \ref{Fig.framework}.
It combines a Mixture of Wavelet Experts (MoWE)–based infrared branch with a generic neural layer-based RGB branch.
MoWE has fewer parameters and higher computational efficiency than generic neural layers.
Compared with conventional symmetric designs, the integration of MoWE substantially enhances the efficiency of asymmetric backbone.

Let $\mathbf{X}_{R}\in \mathbb{R}^{3\times H\times W}$ and $\mathbf{X}_{T}\in \mathbb{R}^{1\times H\times W}$ denote the input tensors for RGB branch and infrared branch, where $H$ and $W$ represent the height and width of input tensors respectively. 
The output multiscale features of RGB branch and infrared branch at $j^{th}$ stage are defined as $\mathbf{I}_R^j\in \mathbb{R}^{C_R^j\times \frac{H}{2^j}\times \frac{W}{2^j}}$ and $\mathbf{I}_T^j \in \mathbb{R}^{C_T^j\times \frac{H}{2^j}\times \frac{W}{2^j}}$, where $j\in \{1,2,3\}$, $C_R^j$ and $C_T^j$ denote the channel numbers of these two branches at $j^{th}$ stage respectively.
Additional two stages are applied to extract higher-level semantic features from the fusion of $\mathbf{I}_R^j$ and $\mathbf{I}_T^j$.
Refined multispectral features from the last three stages are adopted for further pedestrian detection, namely $\mathbf{I}_O^i\in \mathbb{R}^{C^i_O\times \frac{H}{2^i}\times \frac{W}{2^i}}$, where $i\in \{3,4,5\}$ and $C^i_O$ denotes the channel numbers at $i^{th}$ stage.
Residual CNN blocks\cite{he2016deep} are employed in RGB branch and the last two stages.

\subsubsection{Mixture of Wavelet Experts}
\label{Subsec.MoWE}
Infrared imagery generally exhibits weak textures but strong structural and thermal cues, requiring dynamic frequency-domain filtering.
To capture informative infrared patterns with lower computational cost, we utilize Mixture of Wavelet Experts (MoWE) instead of ordinary heavyweight designs.
Given input tensor \(\mathbf{I}_T^{j-1}\in \mathbb{R}^{C_{T}^{j-1}\times \frac{H}{2^{j-1}}\times \frac{W}{2^{j-1}}}\) from $(j-1)^{th}$ stage to $j^{th}$ stage, MoWE leverages multispectral context-aware router to dynamically route different channels of $\mathbf{I}_T^{j-1}$ through sparsely activated ADWT experts and shared DWT experts.
Then subdivided frequency patterns are captured as obtained infrared feature $\mathbf{I}_T^{j}$.
$\mathbf{I}_T^{j}$ is of double downsampled size, and will be subsequently fused with $\mathbf{I}_R^{j}$ to provide complementary details.

\noindent \textbf{Multispectral Context-aware Router.}
To assess environmental context jointly, multispectral context-aware router of MoWE first aggregates information from both current infrared input $\mathbf{I}_T^{j-1}$ and RGB feature $\mathbf{I}_R^{j}$.
Given the number of routed experts $N_e$, routing score $\mathbf{S}_e^{j}\in \mathbb{R}^{c_T^j\times N_e}$ at $j$-th stage is obtained from multispectral context as follows,
\begin{equation}
    \mathbf{S}_e^{j} = \mathbf{W}\cdot\textrm{Flat}(\textrm{Cat}(\mathbb{P}(\mathbf{I}_T^{j-1}),\mathbb{P}(\mathbf{I}_R^{j})))\cdot(1+\lambda_{\epsilon}\epsilon),
\end{equation}

\noindent 
where $\textrm{Cat}(\cdot,\cdot)$ denotes channel concatenation, $\mathbb{P}(\cdot)$ represents adaptive average pooling operation to pool input into $H_p\times W_p$, and $\textrm{Flat}(\cdot)$ denotes flattening operation at last two channels.
$\mathbf{W}\in \mathbb{R}^{N_e\times ((C_{T}^{j-1}+C_R^{j})\cdot H_p\cdot W_p)}$ is learnable weight of MLP layer.
$\epsilon$ is standard Gaussian noise with hyperparameter $\lambda_\epsilon$ to control noise level.
$\epsilon$ enable MoWE to explore diverse routing strategies during training, and is removed during inference.
$\lambda_\epsilon$ is set to 0.2 in this paper.
Then activated experts are selected based on top-$N_{act}$ scores of $\mathbf{S}_e^{j}$ for each channel, resulting in activating score $\mathbf{S}_{\Omega}^{j}\in \mathbb{R}^{C_T^{j-1}\times N_{act}}$, where $N_{act}$ is the number of activated experts for each channel, as follows,
\begin{equation}
    \label{eq.s_omega}
    \mathbf{S}_\Omega^{j} = \operatorname{Softmax}(\{\mathbf{S}_e^{j}[:,k] \mid k \in \Omega\}),\; \Omega = \operatorname{Top-K}(\mathbf{S}_e^{j}),
\end{equation}

\noindent 
where $\operatorname{Top-K}(\cdot)$ denotes top-$N_{act}$ selection operation from $N_{e}$ scores, and $\omega$ is the activated set of experts for each channel.
$\mathbf{S}_e^{j}[:,k]$ means $k^{th}$ column of $\mathbf{S}_e^{j}$.
$\mathbf{S}_\Omega^{j}$ and $\Omega$ are then utilized in subsequent expert aggregation.

\noindent \textbf{Heterogeneous Wavelet Experts.}
Learnable ADWT and static DWT are involved in MoWE.
They can be regarded as the projection onto an orthogonal basis of small finite waves, and both perform 1D convolutions with two types of kernels, \emph{i.e.}, kernel $\mathbf{h}=(h_0,h_1,...,h_K)$ as high-pass filter and kernel $\mathbf{g}=(g_0,g_1,...,g_K)$ as low-pass filter. $K$ denotes kernel length.
The corresponding weights for these filters are initialized from pre-defined wavelet families. 
Notably, there are two differences between ADWT and DWT. 
Firstly, filter kernels $\mathbf{h}$ and $\mathbf{g}$ of ADWT are learnable, and $\mathbf{h}$ is constrained to conjugate mirror filter of $\mathbf{g}$ following \cite{PNASWavelet}.
Secondly, the extracted frequency components of ADWT are further modulated by gain factors.

Besides, for wavelet transform on infrared features, there exist several issues for us to cope with.
Wavelets with long support length are not able to separate close spaced features.
Besides, unsymmetrical wavelets could lead to phase distortion in image processing.
Therefore, symmetrical wavelets with short support length are chosen in this paper. 
Moreover, even-sized kernels of symmetrical wavelets will lead to shrinkage in size and potential shift problem.
Thus, we symmetrically pad $\mathbf{I}_T^{j-1}$ following\cite{wu2019convolution} before convolving $\mathbf{I}_T^{j-1}$ with wavelet kernels.
Then these 1D high-pass and low-pass filters convolve $\mathbf{I}_T^{j-1}$ by stride $s=2$ in both horizontal and vertical directions, resulting in approximation features $\mathbf{A}^{j}$, horizontal detailed features $\mathbf{DH}^{j}$, vertical detailed features $\mathbf{DV}^{j}$, and diagonal detailed $\mathbf{HH}^{j}$.
We assume that $\mathbf{d}^{j}_{:,m,n}$ and $\mathbf{a}^{j}_{:,m,n}$ are obtained by horizontal convolution with $\mathbf{h}$ and $\mathbf{g}$ on $\mathbf{I}_T^{j-1}$ respectively,
\begin{equation}
\left\{\begin{aligned} 
  \mathbf{d}^{j}_{:,m,n}&=\sum\nolimits_{k=1}\nolimits^{K}h_{k}\mathcal{P}(\mathbf{I}_T^{j-1})_{:,m,{sn+k}} \\  
  \mathbf{a}^{j}_{:,m,n}&=\sum\nolimits_{k=1}\nolimits^{K}g_{k}\mathcal{P}(\mathbf{I}_T^{j-1})_{:,m,{sn+k}}
\end{aligned}\right.,
\end{equation}

\noindent 
where $m$ and $n$ denote vertical and horizontal coordinates respectively.
Note that $m\in\{0,1,...,\frac{H}{2^{j-1}}\}$ and $n\in\{0,1,...,\frac{W}{2^{j}}\}$. 
Function $\mathcal{P}(\cdot)$ represents symmetric padding in \cite{wu2019convolution}.
In the same way, infrared frequency features can be obtained by vertical convolution,
\begin{equation}
\left\{\begin{aligned} 
  \mathbf{HH}^{j}_{:,m',n}&=\sum\nolimits_{k=1}\nolimits^{K}h_{k}\mathbf{d}^{j}_{:,sm'+k,n} \\
  \mathbf{DH}^{j}_{:,m',n}&=\sum\nolimits_{k=1}\nolimits^{K}g_{k}\mathbf{d}^{j}_{:,sm'+k,n} \\
  \mathbf{DV}^{j}_{:,m',n}&=\sum\nolimits_{k=1}\nolimits^{K}h_{k}\mathbf{a}^{j}_{:,sm'+k,n} \\  
  \mathbf{A}^{j}_{:,m',n}&=\sum\nolimits_{k=1}\nolimits^{K}g_{k}\mathbf{a}^{j}_{:,sm'+k,n}
\end{aligned} \right., 
\end{equation}

\noindent
where $m'=(0,1,...,\frac{H}{2^{j}})$ and $n=(0,1,...,\frac{W}{2^{j}})$.
Finally, for ADWT, we assign learnable gain factors to modulate frequency components to enhance frequency adaptability to pedestrian detection.
The output of $i$-th ADWT and DWT experts at $j^{th}$ can be respectively represented as follows,
\begin{equation}
\left\{
\begin{aligned}
\mathcal{E}_{act}[i](I_T^{j-1}) &= \boldsymbol{\lambda}^j\cdot \textrm{Cat}(\mathbf{A}^{j},\mathbf{DH}^{j},\mathbf{DV}^{j}, \mathbf{HH}^{j}), \\
\mathcal{E}_{shared}[i](I_T^{j-1}) &= \textrm{Cat}(\mathbf{A}^{j},\mathbf{DH}^{j},\mathbf{DV}^{j}, \mathbf{HH}^{j})
\end{aligned}
\right.
\end{equation}

\noindent
where $\textrm{Cat}(\cdot,\cdot)$ denotes concatenation at channel dimension. 
$\boldsymbol{\lambda}^j\in\mathbb{R}^{C_T^j}$ represents gain factors at $j^{th}$ stage adaptive to datasets, learned by depthwise $1\times 1$ convolution.

\noindent 
\textbf{Expert Aggregation.}
Given $N_s$ shared experts $\mathcal{E}_{shared}$ and $N_{act}$ activated experts $\mathcal{E}_{act}$ of $N_e$ routed experts, the output infrared feature $\mathbf{I}_T^{j}$ at $j^{th}$ stage is as follows,
\begin{small}
\begin{equation}
  \mathbf{I}_T^j = \sum_{i \in \Omega} \textrm{Re}(\mathbf{S}_\Omega^{j}[:,i],4)\cdot\mathcal{E}_{act}[i](\mathbf{X}_T^{j-1}) + \sum_{i=0}^{N_s}\mathcal{E}_{shared} (\mathbf{X}_T^{j-1}),
\end{equation}
\end{small}

\noindent
where $\mathbf{S}_\Omega^{j}[:,i]$ means $i^{th}$ column of $\mathbf{S}_\Omega^{j}$.
$\textrm{Re}(\cdot,4)$ operation repeats the first dimension of $\mathbf{S}_\Omega^{j}[:,i]$ for 4 times.
$\mathbf{I}_T^{j}$ is then fused with RGB feature $\mathbf{I}_R^{j}$ by CMRF module to supplement crossmodal complementary information, and transferred to next stage for further feature extraction.

\noindent 
\textbf{Channel-wise Routing Balance Loss.}
To prevent routing collapse of MoWE during training, following prior work \cite{MoE}, we apply channel-wise routing balance loss $\mathcal{L}_{b}$ to the routing scores $\mathbf{S}_e^{j}$ at three stages.
Let $\eta_{k,j} = \sum_{c=1}^{C_T^{j}}\mathbbm{1}(k \in \Omega[c])$ be routing frequency of $k$-th expert across channel $c$, where $\mathbbm{1}(\cdot)$ denotes the indicator function and  $\Omega[c]$ denotes the subset $\operatorname{Top-K}(\mathbf{S}_e^{j}[c])$ in Eq. \ref{eq.s_omega}. And let $P_{k,j} = \sum_{c=1}^{C_T^{j}}\operatorname{Softmax}(\mathbf{S}_e^{j})[c,k]$ be the average probability of $k$-th expert assigned by the router. $\mathcal{L}_{b}$ is defined as
\begin{equation}
    \mathcal{L}_{b} = \sum_{j=1}^{3} \frac{N_e}{(C_T^{j})^2} \sum_{k=1}^{N_e} \eta_{k,j} \cdot P_{k,j}.
\end{equation}

$\mathcal{L}_{b}$ is added to the total loss with a weight of 0.1 in practice.

\begin{figure}
  \centering
  \includegraphics[width=\textwidth]{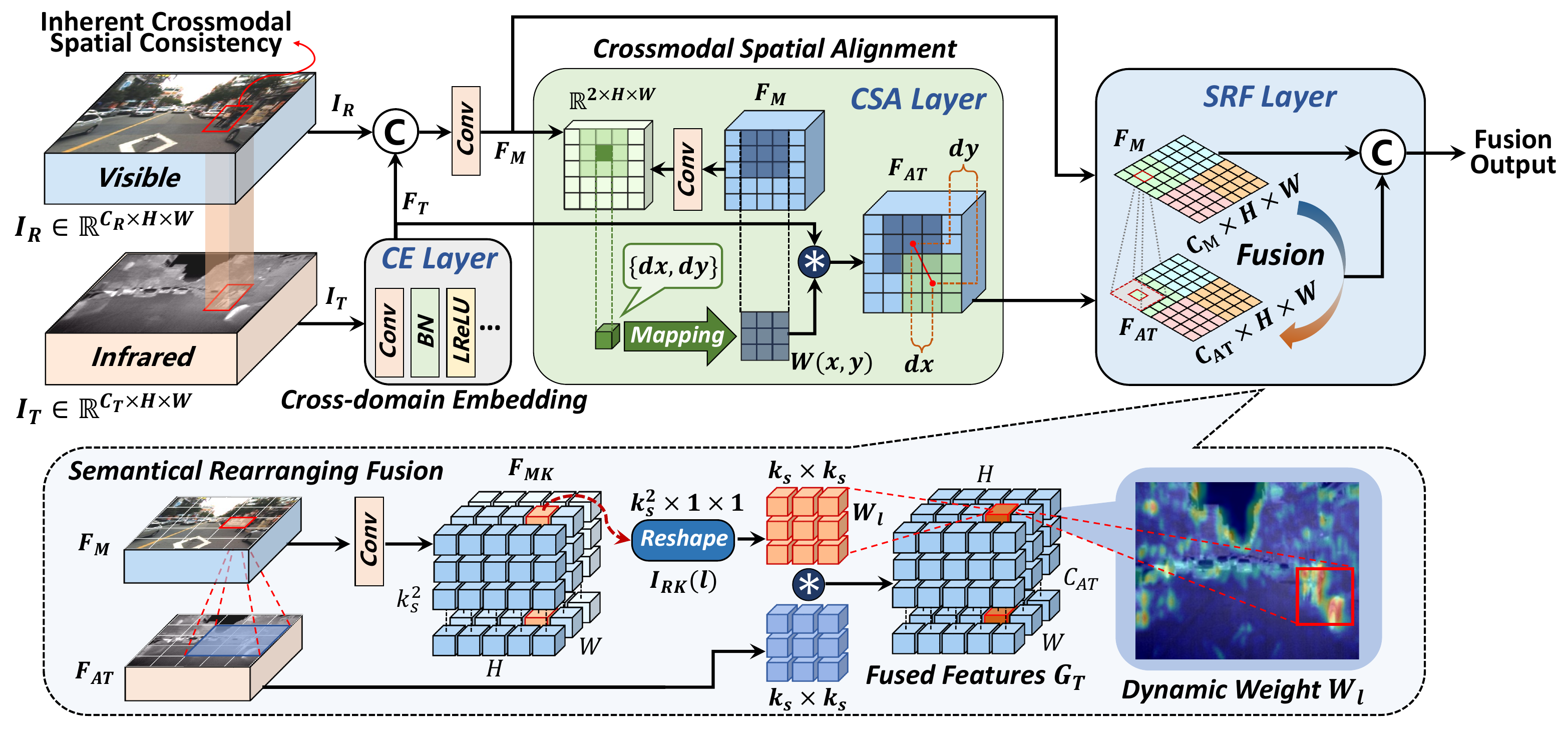}
  \caption{Detailed structure of CMRF. It consists of three parts: Cross-domain Embedding (CE) layer to narrow semantic gap, Crossmodal Spatial Alignment (CSA) layer to maintain spatial consistency, and Semantic Rearranging Fusion (SRF) layer to amplify crossmodal complementary information.\label{Fig. CMRF}  }
\end{figure}

\subsection{Multi-domain Crossmodal Rearranging Fusion\label{Sec: fusion module}}
Multimodal fusion of different modalities is important for accurate multispectral pedestrian detection.
In our framework, RGB features $\mathbf{I}_R^j$ in spatial domain and infrared features $\mathbf{I}_T^j$ in frequency domain have their own characteristic semantics in channel dimension.
The simple amplification for numerically different features in existing approaches \cite{MBNet2020,tang2022piafusion,wang2022RISNet,zhang2023TripleNet} may ignore the potential semantic gap between those heterogeneous features of different spectra, and turns out to be nearly ineffective as demonstrated in Sec. \ref{Exp.ablation}. 
In spite of semantic diversity, there exists inherent spatial consistency cross multimodal feature maps.
In light of this, we propose a novel fusion module named Crossmodal Rearranging Fusion (CMRF).
As illustrated in Fig. \ref{Fig. CMRF}, CMRF consists of three parts: Cross-domain Embedding (CE) layer, Crossmodal Spatial Alignment (CSA) layer, and Semantic Rearranging Fusion (SRF) layer.

\subsubsection{Cross-domain Embedding}
At the $j^{th}$ stage, infrared features $\mathbf{I}_T^{j}$ contains frequency-domain components which is complementary but contrasting to RGB features $\mathbf{I}_R^{j}$.
Therefore, Cross-domain Embedding Layer (CE) are adopted to embed $\mathbf{I}_T^{j}$ into the same domain of features in RGB branch.  
The embedding layer consists of a few embedding units, specifically two units in our implementation, which conduct $3\times 3$ convolution, batch normalization, and activation function in sequence.
Taking one embedding unit as example, for input tensor $\mathbf{I}_T^{j}$, the embedding unit generates the output tensor $\mathbf{F}_T^j\in \mathbb{R}^{C_T^j\times \frac{H}{2^j}\times \frac{W}{2^j}}$ of the same size, 
\begin{equation}
\label{Eq.encoder}
\mathbf{F}_T^j = \mathrm{LReLU}\left (\alpha\cdot\frac{\mathcal{F}_{conv}(\mathbf{I}_T^{j})-\mathrm{E}(\mathcal{F}_{conv}(\mathbf{I}_T^{j}))}{\sigma(\mathcal{F}_{conv}(\mathbf{I}_T^{j}))}+\beta\right ),
\end{equation}

\noindent 
where $\alpha$ and $\beta$ denote learnable affine parameters, and function $\mathcal{F}_{conv}(\cdot)$ denotes the convolution operation with learnable weights. $\mathrm{E}(\cdot)$ and $\sigma(\cdot)$ denote the mean and the standard deviation respectively. Leaky ReLU (LReLU) is adopted as activation. 
% FIXME: LReLU \cite{maas2013rectifier} 
In this way, for further crossmodal fusion, the $j^{th}$-stage CE layer projects $\mathbf{I}_T^{j}$ into semantically embedded infrared features $\mathbf{F}_T^j$. 

\begin{figure}
  \centering
  \subfloat[KAIST Training Set]{\includegraphics[width=0.4\textwidth]{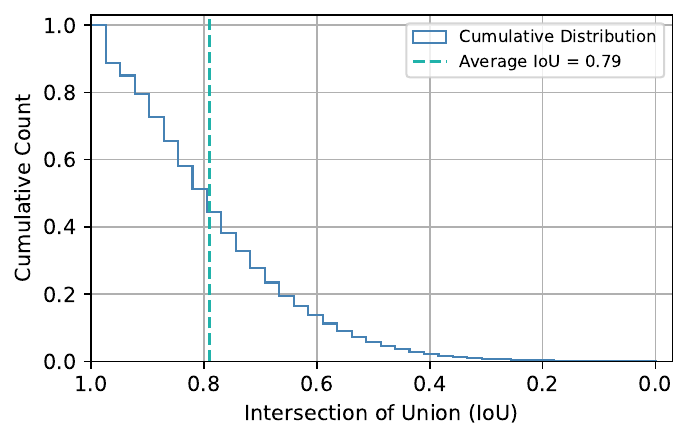}} 
  \subfloat[KAIST Sanitized Test Set]{\includegraphics[width=0.4\textwidth]{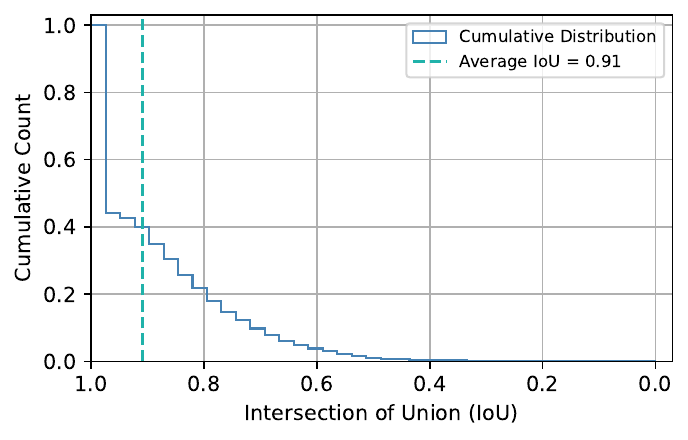}}  
  \caption{The statistic of Intersection over Union (IoU) of bounding boxes between paired RGB and infrared images in (a) KAIST training set and (b) KAIST test set with sanitized annotations.}
  \label{fig:misalign} 
\end{figure} 

\subsubsection{Crossmodal Spatial Alignment}
Since multispectral image pairs are often weakly aligned, it is vital to align feature maps of two spectra to assure crossmodal spatial consistency for subsequent crossmodal fusion.
Inspired by \cite{dai2017deformable}, we propose Crossmodal Spatial Alignment layer to predict pixel-wise offsets of infrared modality relative to RGB modality for alignment. 
Differing from self-shifting in \cite{dai2017deformable}, the offsets in our framework are predicted from features of different modalities simultaneously and utilized for crossmodal shifting.

Firstly, at $j^{th}$ stage, RGB features $\mathbf{I}_R^j$ and semantically embedded infrared features $\mathbf{F}_T^j$ are concatenated in channel dimension and then convolved by a $3\times 3$ convolution layer to get the multimodal representation, \emph{i.e.}, $\mathbf{F}_{M}^j\in \mathbb{R}^{C_M^j\times \frac{H}{2^j}\times \frac{W}{2^j}}$.
From this representation, another $3\times 3$ convolution layer is applied to predict pixel-wise offsets. 
The predicted offset map has the same spatial resolution with $\mathbf{F}_{M}^j$, in which channel dimension of pixel $(m,n)$ is 2 corresponding to 2D offsets $\Delta \mathbf{P}(m,n) = (\dif x,\dif y)$.
To smoothly rearrange the pixels of $\mathbf{F}_T^j$, pixels of $\mathbf{F}_T^j$ are not directly shifted by offsets and interpolated to get a spatial aligned feature map $\mathbf{F}_{AT}^j$.
Instead, the pixels of $\mathbf{F}_{AT}^j$ are the weighted sums of corresponding shifted receptive field of $\mathbf{F}_T^j$, which can be written as
\begin{small}
  \begin{equation}
    \mathbf{F}_{AT}^j(m,n) =\! \sum_{x=-w}^{+w} \sum_{y=-h}^{+h} \mathbf{U}(x,y) \mathbf{F}_T^j(m+x+\dif x,n+y+\dif y),
  \end{equation}
\end{small}

\noindent
where the output feature map $\mathbf{F}_{AT}^j$ is of the same size with $\mathbf{F}_T^j$. $\mathbf{U}$ denotes learnable weights of $K \times K$ odd size centered at (0,0), $w$ and $h$ equal to $\frac{K-1}{2}$. $m$ and $n$ are the 2D coordinates in feature maps. 
Since $\dif x$ and $\dif y$ are fractional, the pixels of coordinate $(m+x+\dif x,n+y+\dif y)$ in $\mathbf{F}_T^j$ are approximated via bilinear interpolation of neighboring pixels, namely pixels at $(m+\lfloor x+\dif x \rfloor,n+\lfloor y+\dif y\rfloor)$, $(m+\lceil x+\dif x \rceil,n+\lfloor y+\dif y\rfloor)$, $(m+\lfloor x+\dif x \rfloor,n+\lceil y+\dif y \rceil)$, and $(m+\lceil x+\dif x \rceil,n+\lceil y+\dif y \rceil)$.

\subsubsection{Semantic Rearranging Fusion}
Despite the features of infrared modality are accommodated to RGB modality so far, the features further need to adaptively amplify complementary information of pedestrians which are semantically similar across different modalities.
Instead of numerically complementary feature extraction in prior works, we propose Semantic Rearranging Fusion (SRF) layer for semantic-aware fusion.
SRF leverages inherent crossmodal spatial consistency and transfers semantically complementary infrared features in relevant local regions to complement the input multimodal representation, further recalling reliable pedestrian features from infrared modality.

We first briefly explain the overall process. 
Recall that multimodal representation $\mathbf{F}_{M}^j\in \mathbb{R}^{C_M^j\times \frac{H}{2^j}\times \frac{W}{2^j}}$ contain combined pedestrian patterns, and aligned infrared features $\mathbf{F}_{AT}^j\in \mathbb{R}^{C_T^j\times \frac{H}{2^j}\times \frac{W}{2^j}}$ are derived from CSA layer as the input multimodal tensors for SRF.
Following crossmodal spatial consistency, SRF exploits the relationship between channel-dimension semantic contents of the pixel $\mathbf{F}_M^j(l)$ located at $l=(m,n)$ of $\mathbf{F}_M^j$ and the corresponding local region centered at $l=(m,n)$ in $\mathbf{F}_{AT}^j$. 
The channel-dimension semantic contents of $\mathbf{F}_M^j(l)$ are utilized to generate the corresponding local rearranging weight $\mathbf{W}_l$. 
Then $\mathbf{W}_l$ is applied to extract instance-specific complementary features from $\mathbf{F}_{AT}^j$ through calculating the weighted sum for pixels of $\mathbf{F}_{AT}^j$ spatially relevant to $\mathbf{F}_M^j$, namely the pixels of corresponding local region centered at $l$ in $\mathbf{F}_{AT}^j$. 

To be more specific, convolution layers are first performed on $\mathbf{F}_M^j$ to compress the channel number $C_M^j$ into a smaller one $C_{MK}^j$, resulting in $\mathbf{F}_{MK}^j\in \mathbb{R}^{C_{MK}^j\times \frac{H}{2^j}\times \frac{H}{2^j}}$.
Then given $C_{MK}^j=k_r^2$ where $k_r\in \mathbb{N}^*$, the pixel $\mathbf{F}_{MK}^j(l)\in \mathbb{R}^{k_r^2\times 1\times 1}$ located at $l=(m,n)$ is spatially reshaped to get the local rearranging weight $\mathbf{W}_l\in \mathbb{R}^{k_r\times k_r}$. 
The obtained $\mathbf{W}_l$ is spatially normalized by softmax function to avoid altering the mean of the corresponding region in $\mathbf{F}_{AT}^j$, which sets the sum of elements in $\mathbf{W}_l$ to $1$. 
The generation of $\mathbf{W}_l$ mentioned above can be formally written as
\begin{equation}
\mathbf{W}_l=\textrm{Softmax}\left(\mathcal{F}_{conv}(\mathbf{F}_{M}^j(l))\right),
\end{equation}

\noindent 
where $\mathcal{F}_{conv}$ represents convolution operations with learnable weights. 
$\mathbf{W}_l$ generated from $\mathbf{I}_R^j(l)$ will focus on $\mathbf{F}_M^j(l)$ if $\mathbf{F}_M^j(l)$ suspected to contain pedestrian instances.
% , which is illustrated by the visualization of $\mathbf{W}_l$ in Sec. \ref{Exp.ablation}.

At last, instance-specific complementary features $\mathbf{G}_{T}^j$ of infrared modality are extracted by convolving the corresponding local region in $\mathbf{F}_{AT}^j$ with $\mathbf{W}_l$ centered at $(m,n)$,
\begin{equation}
\mathbf{G}_{T}^j(l) = \sum_{x=-k}^{k} \sum_{y=-k}^{k} \mathbf{W}_l(x,y)\mathbf{F}_{AT}^j(m+x,n+y),
\end{equation}
where $k=\lfloor\frac{k_r}{2}\rfloor$. 
At last, $\mathbf{G}_{T}^j$ is concatenated with $\mathbf{F}_M^j$ as next stage input, semantically complementing pedestrian features, which is visualized in Sec. \ref{Exp.ablation}.

\section{Experiments}
% In this section, comprehensive evaluation of WCCNet is evaluated on several important multispectral pedestrian detection benchmarks and compared with other state-of-the-art methods. 
% Comprehensive ablation experiments are also conducted with quantitative and qualitative illustrations.

\subsection{Benchmarks\label{Exp.datasets}} 
Our approach is evaluated on two widely-used multispectral pedestrian detection benchmarks, i.e., KAIST and FLIR.

KAIST \cite{KAIST2015} contains videos of 95,328 RGB-infrared paired frames with 103,128 bounding boxes in total,
which are taken from urban road scenes during day and night to cover changes in diverse lighting conditions, heavy occlusions, and multiple pedestrian scales. 
The videos are split into two halves following \cite{KAIST2015}, where the first half is used for training, and the latter half for testing. 
Then those two set of videos are sampled every 20 frames into images to obtain image datasets for training and testing respectively.
Among images for training, $20$ percent of them are randomly selected for validation in training phase.
Following common annotation settings for a fair comparison with recent related works, the paired annotations in \cite{ARCNN2019} are adopted for training set and the sanitized annotations \cite{liu2016CleanKAIST} are applied for testing set.

FLIR \cite{FLIRDataset} contains targets of richer categories than KAIST, including person, car and bicycle. 
RGB and infrared image pairs in FLIR are taken from moving vehicles at various situations. 
Following common annotation settings \cite{CFR2020}, misaligned RGB-infrared image pairs are removed, resulting in 4,129 image pairs for training and 1,013 image pairs for testing.

\subsection{Implementation Details\label{Exp.details}}

\subsubsection{Platform and Data Processing}
The proposed WCCNet has extended the framework of YOLOX \cite{ge2021yolox} in PyTorch. 
We utilize NVIDIA RTX 3090 GPUs for training, and employ a series of GPUs for testing, as shown in Table \ref{Tab. KAIST}. 
For preprocessing, RGB images and infrared images are all resized to $640\times 640$ before being input to network. 
For post-processing, the threshold of Non-Maximum Suppression (NMS) is $0.65$, and the threshold for confidence score of positive targets is set to $0.01$.
We split training process into two stages, namely pretraining and supervised training.

\begin{figure}[!h]
  \centering
  \includegraphics[width=0.6\textwidth]{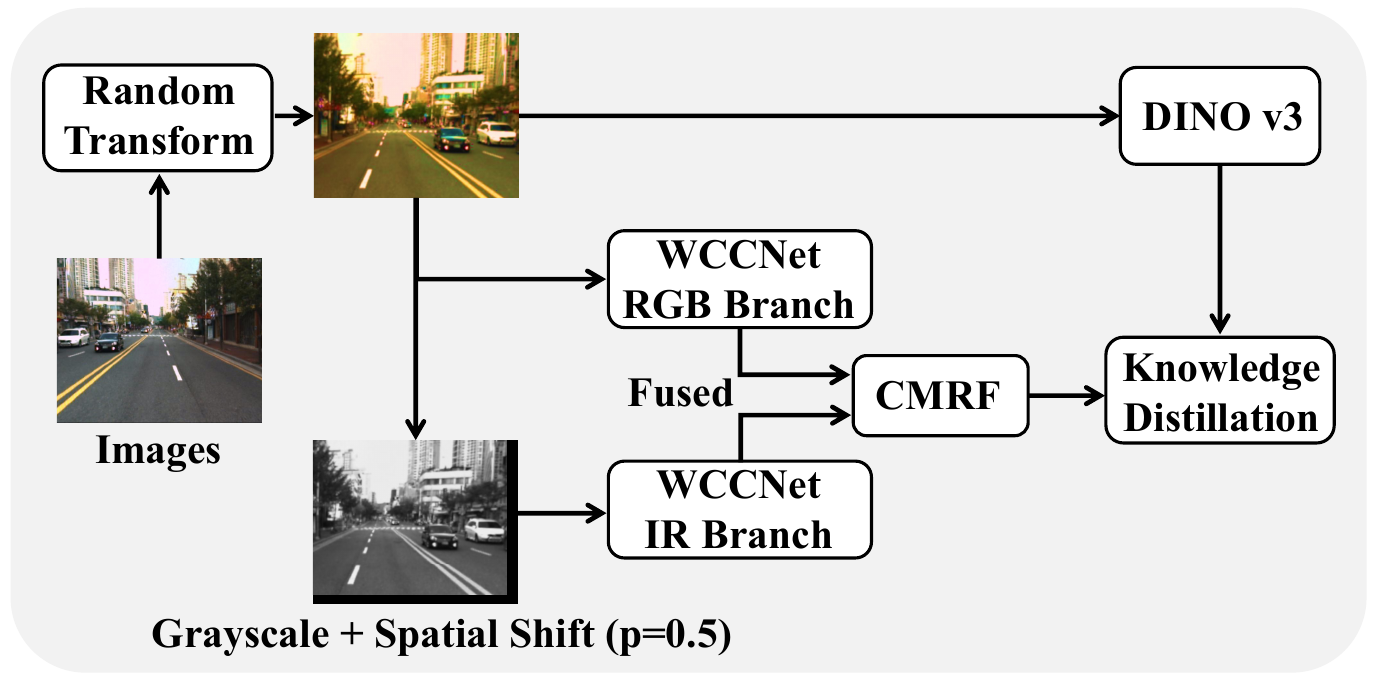}
  \caption{Pretraining strategy of WCCNet backbone. Elaborated data augmentation operations are applied.\label{fig. pretrain}}
  \vspace{-20pt}
\end{figure}

\subsubsection{Pretraining Setting}
The backbone of WCCNet is first pretrained on a union dataset to be distilled from DINOv3-ConvNeXt-Base \cite{dinov3} model following \cite{beyer2022knowledge}, as shown in Fig. \ref{fig. pretrain}.
The union dataset combines COCO-Minitrain-25k dataset \cite{COCO2014} and 50\% images with Top-50\% average target size from KAIST training set.
WCCNet backbone is trained by AdamW optimizer for 50 epochs with initial learning rate of 1e-3 and total batch size of 64.
Extracted multiscale features from CMRF modules in WCCNet backbone is constrained to be consistent with the ones of DINOv3 by MSE loss, and EMA strategy is adopted.

Images first undertakes random transformation, including random cropping, rotation, color jittering, and etc.
Then augmented images are sent to WCCNet and DINOv3 respectively.
Notably, for infrared branch of WCCNet, images are converted to grayscale and have 50\% probability to be randomly shifted to simulate misalignment.
The shift range is set to $\pm 0.1W$ pixels, where $W$ denotes the width of feature maps, and the blank area caused by shifting is padded with zeros.
\begin{table}[!t]
  \centering
  \resizebox{\linewidth}{!}{
  \begin{tabular}{ccccccccccc}
  \toprule
  \multirow{2}{*}{Methods}& \multicolumn{4}{c}{Time Complexity \& Space Complexity}& \multicolumn{3}{c}{$\mathrm{MR}^{-2}$ (All)}& \multicolumn{3}{l}{$\mathrm{MR}^{-2}$ (Reasonable)}\\
  \cmidrule(l){2-5} \cmidrule(l){6-8} \cmidrule(l){9-11}
  & Params.(M)& FLOPs(G)&Time(ms)& GPU & All& Day& Night& All& Day& night
  \\ \midrule 
  Fusion RPN+BF\cite{RPN2017Konig} & $>138.4$  & $>126.20$  & 800& -  & -  & -  & -  & 29.83  & 30.51  & 27.62  \\
  MSDS\cite{MSDS2018}  & $>71.56$ & $>480.03$ & 228 & TITAN X & 34.20   & 32.12  & 38.83  & 11.34  & 10.54  & 12.94  \\
  ARCNN\cite{ARCNN2019} & $>43.88$ & $>353.79$ & 120& 1080TI & 34.95  & 34.36  & 36.12  & 9.34   & 9.94   & 8.38   \\
  CIAN\cite{CIAN2019} & $>162.5$ & $>142.19$  & 66.67  & 1080TI & 35.57  & 36.06  & 32.38  & 14.12  & 14.78  & 11.13  \\
  TC-DET\cite{TCDET2020} & $>63.00$  & $>156.13$  & 33.17  & 1080   & -  & -  & -  & 27.11  & 34.81  & 10.31  \\
  MBNet\cite{MBNet2020}& $>99.23$  & $>139.96$  & 70 & 1080TI & 31.87  & 32.39  & 30.95  & 8.13   & 8.28   & 7.86   \\
  MLPD\cite{MLPD2021}  & $>162.5$  & $>142.19$  & 12 & 2080TI & 28.49  & 28.39  & 28.69  & 7.58   & 7.96   & 6.95   \\
  RISNet\cite{wang2022RISNet}  & -  & -  & -  & -  & -  & -  & -  & 7.89   & \textcolor{blue}{7.61}   & 7.08   \\
  BAANet\cite{yang2022BAANet} & $>73.96$  & -  & 70 & 1080TI & -  & -  & -  & 7.92   & 8.37   & 6.98   \\
  CMPD\cite{CMPD2022}  & $>99.23$  & $>139.96$  & 110& 1080TI & 28.98  & 28.30   & 30.56  & 8.16   & 8.77   & 7.31   \\
  GAFF\cite{GAFF2021} & 31.43 & - & \textcolor{green}{9.34} & 1080TI & 27.30  & 30.59  & 19.22  & \textcolor{blue}{6.48}   & 8.35  & \textcolor{blue}{3.46} \\
  TINet\cite{zhang2023TripleNet} & 61.00 & -  & 87.7   & TITAN X& -  & -  & -  & 9.15   & 10.25  & 7.48 \\
  TFNet\cite{TFNet2024} & - & -  & 150   & TITAN X & -  & -  & -  & 8.09   & 8.41  & 7.34 \\
  ICAFusion\cite{ICAFusion} & 120.2 & - & 26 & 3090 & -  & -  & -  & 7.17   & \textcolor{red}{6.82}  & 7.85 \\ 
  MS-DETR\cite{MS-DETR2024} & 76.38 & 258.84 & 110 & 3090 & 20.64  & 22.76  & 15.77  & \textcolor{red}{6.13}   & \textcolor{green}{7.78}  & \textcolor{red}{3.18} \\
  \cmidrule{1-11}
  \multirow{3}{*}{WCCNet} & \multirow{3}{*}{\textcolor{green}{17.50}}  & \multirow{3}{*}{\textcolor{green}{91.00}}  & 10.62 & 3090   & \multirow{3}{*}{\textcolor{red}{17.81}}  & \multirow{3}{*}{\textcolor{red}{20.36}}  & \multirow{3}{*}{\textcolor{red}{12.37}}  & \multirow{3}{*}{\textcolor{green}{6.72}}   & \multirow{3}{*}{\textcolor{black}{7.80}}   & \multirow{3}{*}{\textcolor{green}{4.72}} \\
    &   &  & 19.08 & 1080TI   &   &   &  &  &  & \\
    &   &  & 36.43 & TITAN X   &   &   &  &  &  & \\
  \cmidrule{1-11}
  WCCNet-S & \textcolor{blue}{9.75}  & \textcolor{blue}{53.86}  & \textcolor{blue}{6.56}   & 3090   & \textcolor{blue}{19.54}  & \textcolor{blue}{21.77}  & \textcolor{blue}{14.39}  & 7.30   & 8.53   & 5.03   \\
  WCCNet-XS & \textcolor{red}{7.78}  & \textcolor{red}{43.87}  & \textcolor{red}{4.45}   & 3090   & \textcolor{green}{20.57}  & \textcolor{green}{22.61}  & \textcolor{green}{15.68}  & 8.92   & 10.06   & 6.62   \\
  \bottomrule
  \end{tabular}}
  \caption{Comparisons with the state-of-the-art methods on KAIST dataset. - means not reported in the corresponding paper. $>$ means the estimated lower bound for the metrics not reported in the corresponding paper. \textcolor{red}{Red}, \textcolor{blue}{blue}, and \textcolor{green}{green} colors denote the first, the second, and the third place respectively.\label{Tab. KAIST}}
\end{table}

\subsubsection{Supervised Training Setting}
WCCNet is further trained on training sets of involved benchmarks.
Pretrained backbone weight is loaded for initialization.
AdamW optimizer is adopted for 80 epochs with initial learning rate of 1e-3 and total batch size of 32.
Mosaic and MixUp data augmentations are adopted for the first three quarters of training epochs.
% Other hyperparameters follow YOLOX. 
% Mosaic \cite{bochkovskiy2020yolov4} 
% Mixup \cite{zhang2018mixup}

\begin{figure}[!t]
    \centering
    % 设置子图标题字体大小
    \captionsetup[subfloat]{font=scriptsize}

    % =============================================
    % 第一行: All-dataset setting
    % =============================================
    \begin{minipage}[c]{0.04\linewidth}
        \centering
        % rotatebox需要 graphicx 宏包
        \rotatebox{90}{\textbf{All-dataset setting}} 
    \end{minipage}%
    \hfill
    \begin{minipage}[c]{0.95\linewidth}
        \centering
        \subfloat[All Time]{
            \includegraphics[width=0.3\linewidth]{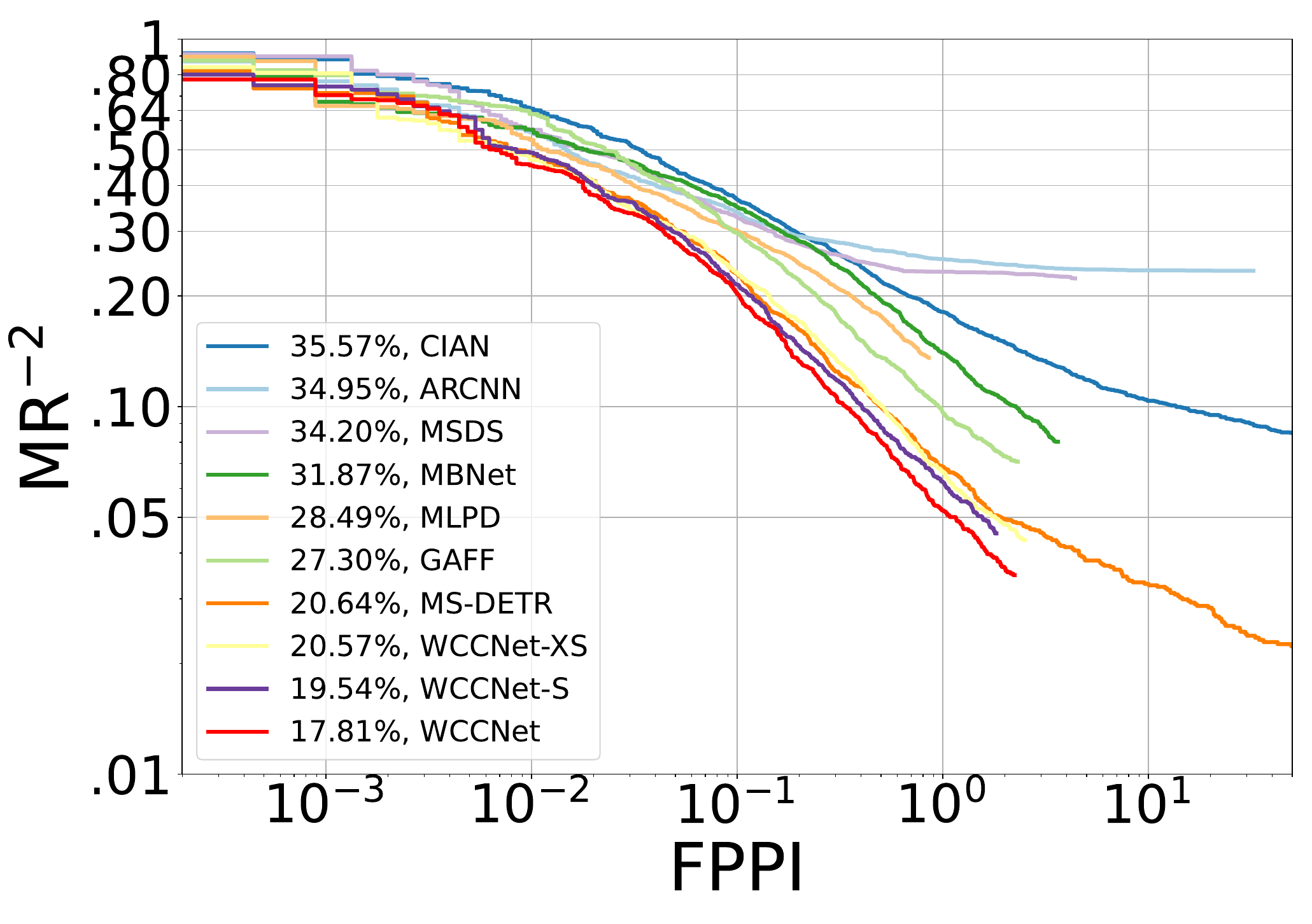}
        }
        \hfill
        \subfloat[Daytime]{
            \includegraphics[width=0.3\linewidth]{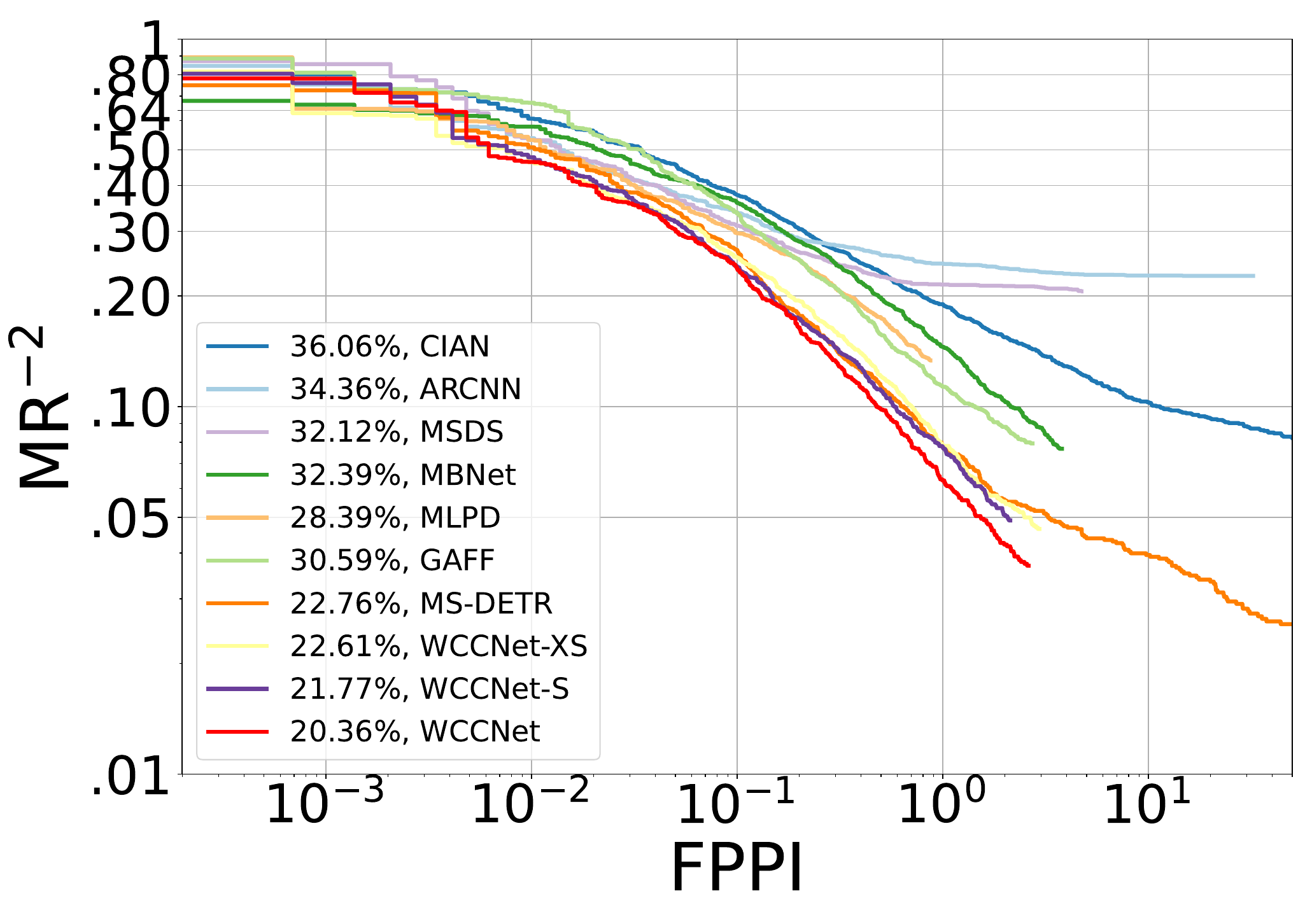}
        }
        \hfill
        \subfloat[Night]{
            \includegraphics[width=0.3\linewidth]{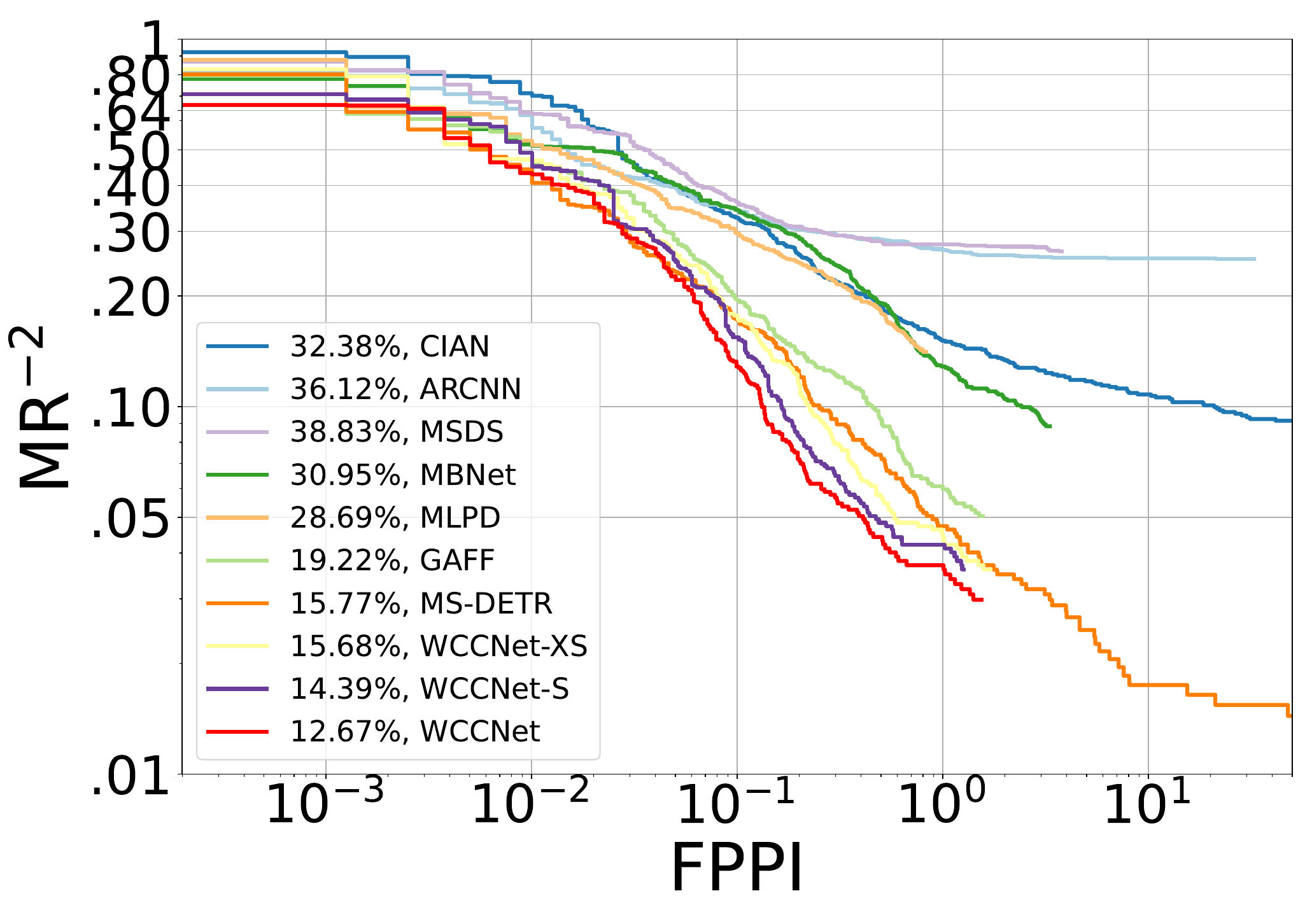}
        }
    \end{minipage}

    \vspace{2mm} % 两行大图之间的间距

    % =============================================
    % 第二行: Reasonable setting
    % =============================================
    \begin{minipage}[c]{0.04\linewidth}
        \centering
        \rotatebox{90}{\textbf{Reasonable setting}}
    \end{minipage}%
    \hfill
    \begin{minipage}[c]{0.95\linewidth}
        \centering
        \subfloat[All Time]{
            \includegraphics[width=0.3\linewidth]{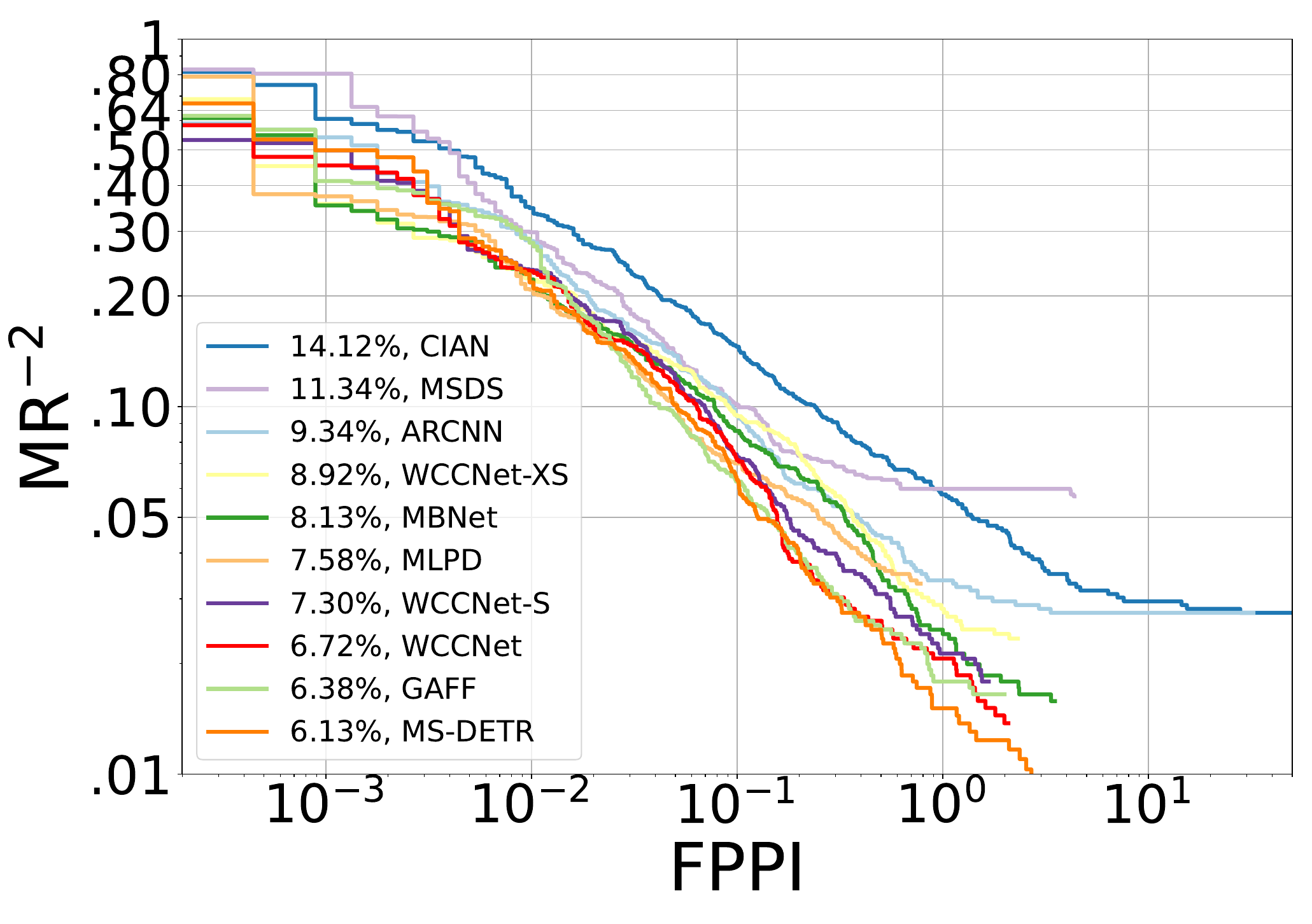}
        }
        \hfill
        \subfloat[Daytime]{
            \includegraphics[width=0.3\linewidth]{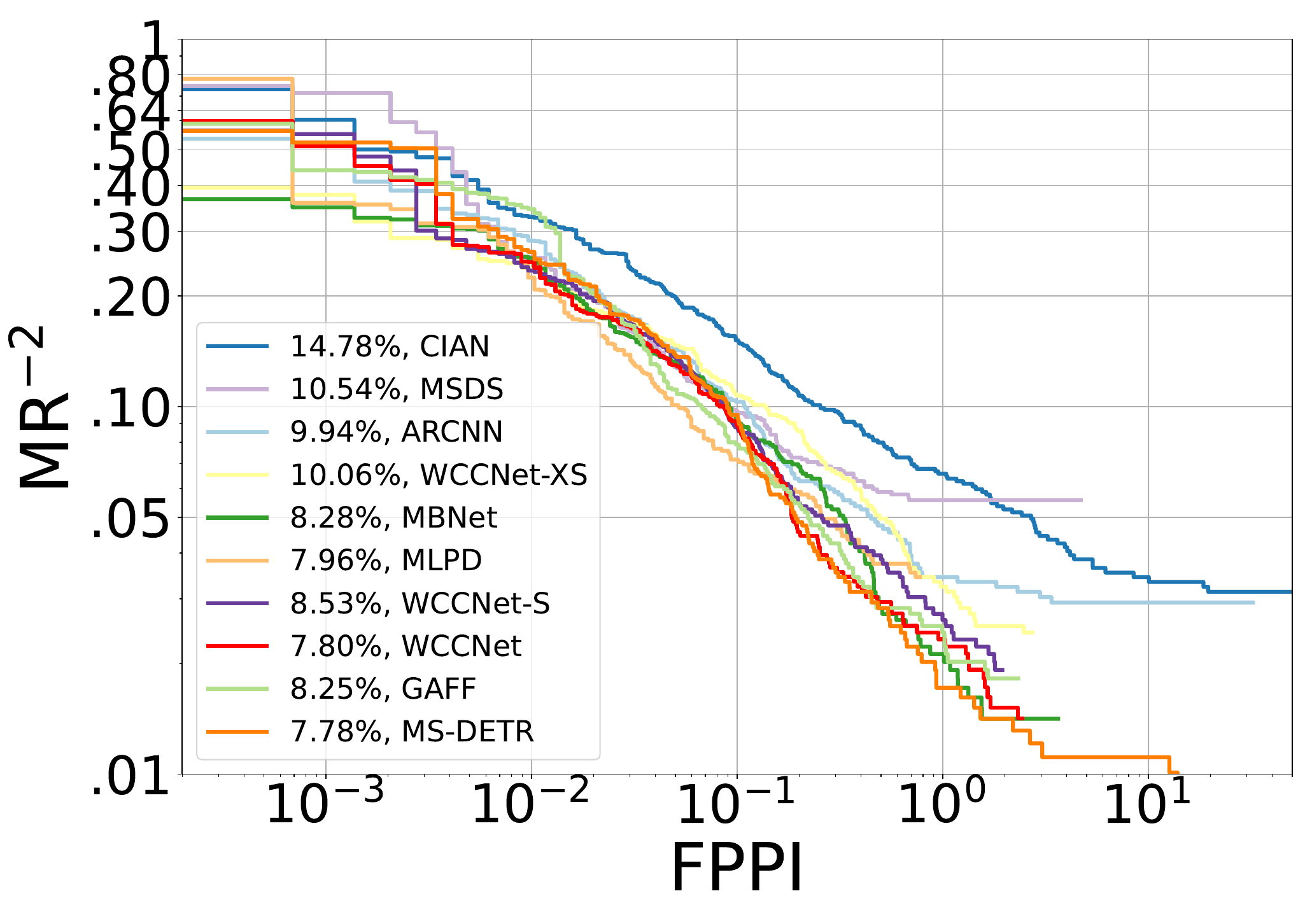}
        }
        \hfill
        \subfloat[Night]{
            \includegraphics[width=0.3\linewidth]{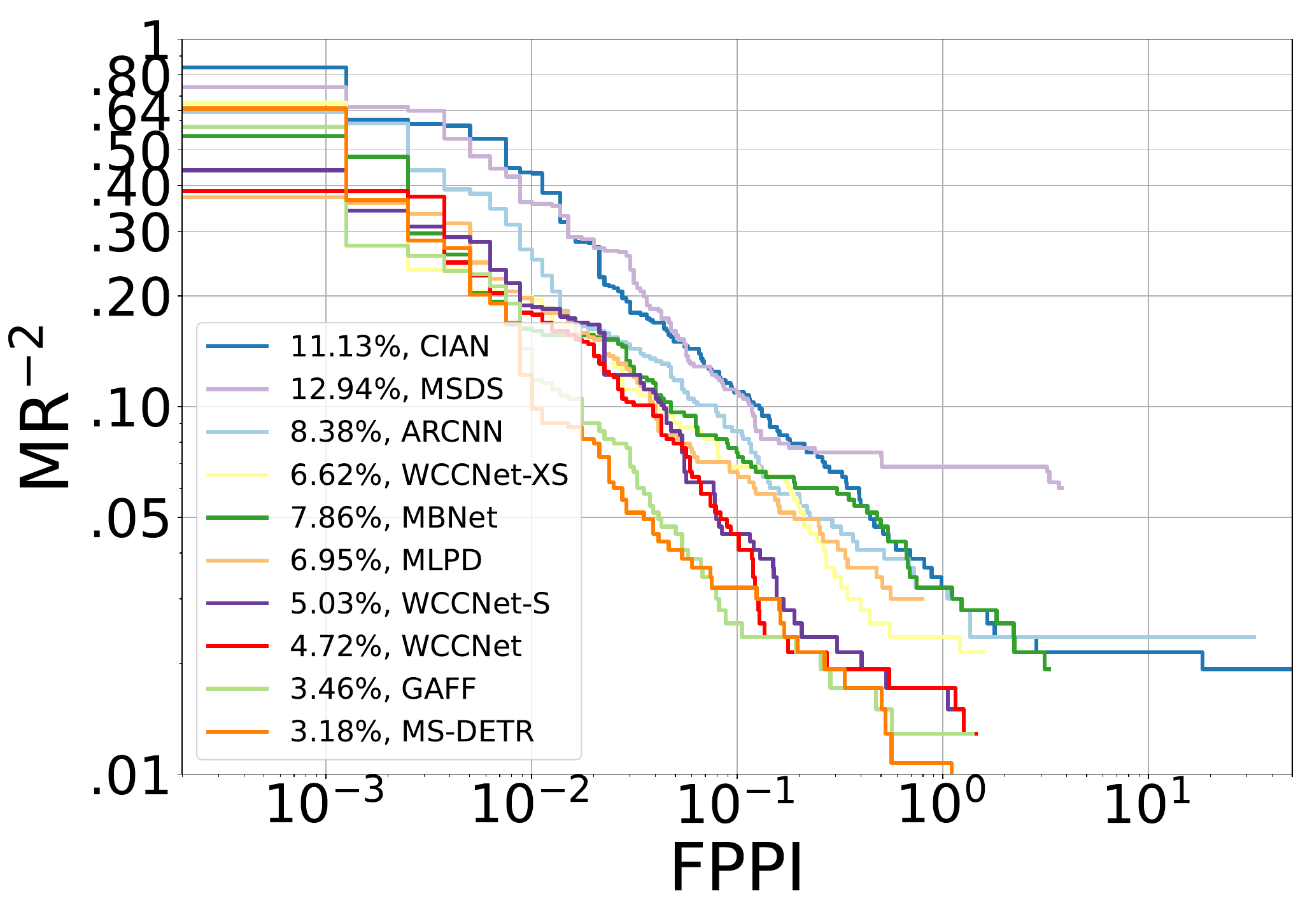}
        }
    \end{minipage}

    \caption{Log-scale MR-FPPI curves on KAIST with different experimental settings and different illumination conditions. The lower the curve, the higher the accuracy represented. \label{Fig. KAIST_Curve}}
\end{figure}

\subsubsection{Supplementary Details for Network}
A lightweight convolutional network is adopted as RGB-branch backbone, which follows DarkNet \cite{redmon2018yolov3} family and has 34 convolutional layers.
It has five stages with $\{1,2,4,3,2\}$ residual blocks respectively.
The first three stages are applied in RGB branch for feature extraction and the last two stages are shared by two modalities for further multispectral detection.
The number of channels $C_R^j$ at $j^{th}$ stage is adjustable with a scaling parameter $\tau$ and set to $2^j \times 16\tau$, while the number of channels for $j^{th}$-stage embedded feature maps $F_T^j$ in CE layer of infrared branch is set to $2^j \times 8\tau$. 
In our implementation, the $\tau$ of WCCNet is set to $1.0$.
Besides, for comprehensive analysis on the impacts of different amounts of parameters, we also present two more efficient types of WCCNet, namely WCCNet-S with $\tau=0.5$, and WCCNet-XS with $\tau=0.25$. 
Smaller $\tau$ means lower computational cost but greater underfitting risk, resulting in faster inference speed but lower accuracy.
% WCCNet with $\tau=1.0$ is recommended in practical applications for its high accuracy and sufficient speed.

For MoWE, routed expert number $N_e$ is set to 8, activated expert number $N_{act}$ is set to 4, and shared expert number $N_s$ is set to 2.
Haar and db3 wavelet are selected for static DWT experts, and the support length $K$ of ADWT experts is set to 6. 
Pooling size for MoWE router is set to $4\times 4$.

For the neck of WCCNet, feature pyramid mechanisms are adopted, including Spatial Pyramid Pooling (SPP) \cite{SPP2015He} and Feature Pyramid Network (FPN) \cite{lin2017FPN}.
Specifically, SPP is applied after the last stage of the backbone for further extraction of high-level semantic information.
Besides, FPN is applied for top-down information interaction between features of different scales.
Detection head for classification and regression of bounding boxes are following YOLOX \cite{ge2021yolox}.

\begin{figure}[!t]
  \centering
  \includegraphics[width=0.9\textwidth]{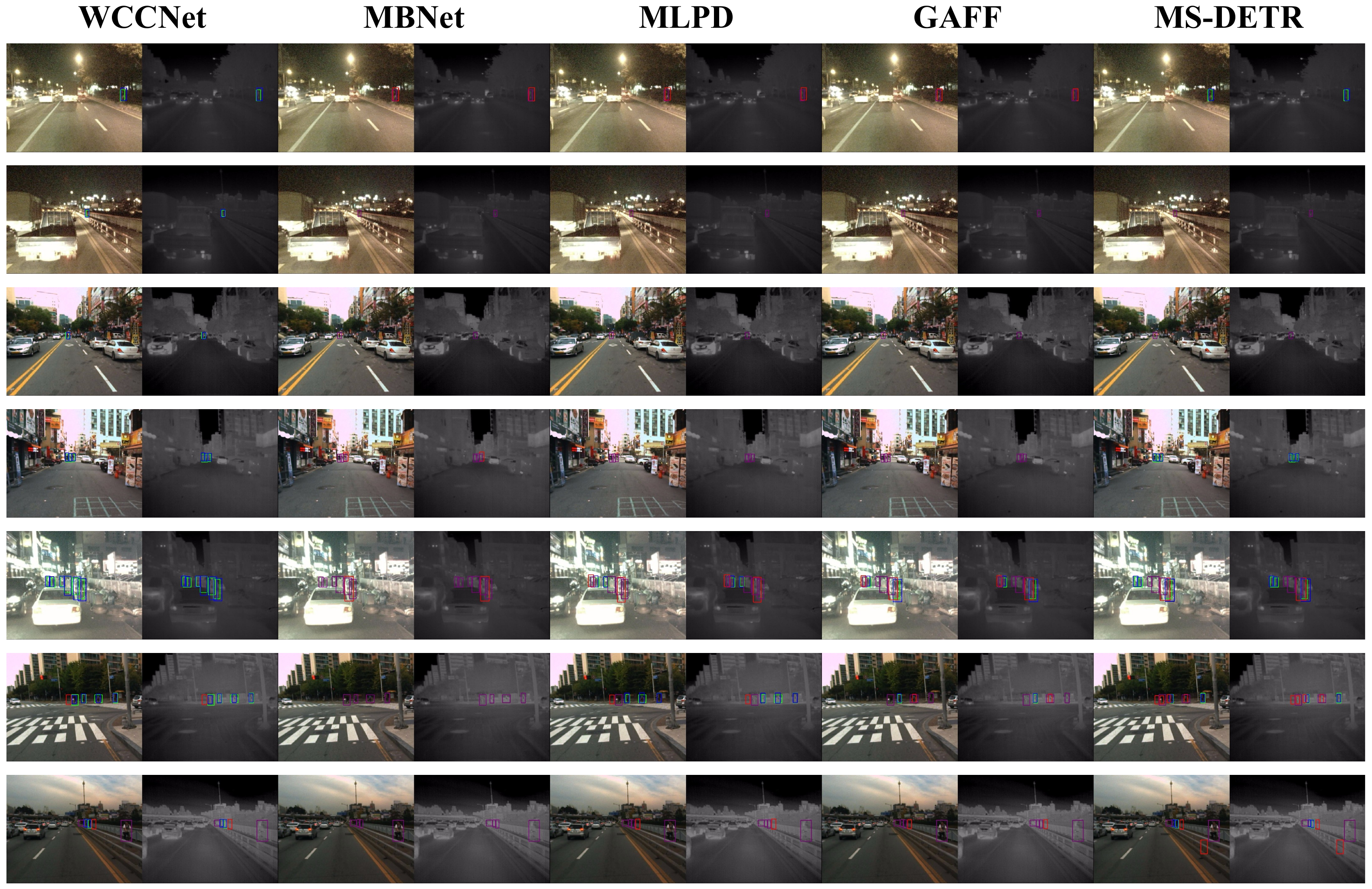}
  \caption{Comparison on detection results with prior works\cite{MBNet2020,MLPD2021,GAFF2021,MS-DETR2024}, which are overlaid on RGB/infrared image pairs in two columns. Difficult scenarios are involved, including occlusions, interfering targets, and small-scale pedestrians. \textcolor{blue}{Blue}, \textcolor{red}{red} and \textcolor{purple}{purple} bounding boxes stand for true positives, false positives, and false negatives, respectively. Corresponding ground truth is shown by \textcolor{green}{green} box for each true positive predictions. \label{Fig. det_result}}
\end{figure}

\subsection{Quantitative Metrics}
The metric $\mathrm{MR}^{-2}$ is used for evaluating the accuracy on KAIST.
It equals to the area of log-average Miss Rate (MR) over False Positive Per Image (FPPI), where FPPI ranges from $10^{-2}$ to $1$. 
Besides, the log-average MR-FPPI curves are obtained by varying the threshold of detector from $0$ to $1$ to generate different MR-FPPI points and plotting these points in logarithmic-scale axes.
For consistent comparison with previous works on FLIR, the COCO-style \cite{COCO2014} evaluation metrics are adopted, namely Average Precision (AP) for specific class and mean Average Precision (mAP) for all classes. 
To evaluate the efficiency of multispectral pedestrian detectors, the memory cost are denoted by the numbers of parameters (Param.), while computational complexity are indicated by the number of floating point operations (FLOPs) and the inference time (Time). 
Notably, the inference time of our model includes post-processing time.
Many previous works may not prioritize the efficiency of their models and have not report corresponding metrics in their papers, thus we estimate their lower bound.
The FLOPs metric is estimated based on $640\times 640$ input size.

\begin{table}[!t]
  \centering
  \resizebox{0.6\textwidth}{!}{
  \renewcommand{\arraystretch}{0.9}
  \begin{tabular}{cccccc}
    \toprule
    \multicolumn{1}{c}{\multirow{2}{*}{Method}} & \multicolumn{1}{c}{\multirow{2}{*}{Time (ms)}} & \multicolumn{1}{c}{\multirow{2}{*}{mAP50}} & \multicolumn{3}{c}{AP50 per class}\\
    \cmidrule{4-6}
    \multicolumn{1}{c}{} & \multicolumn{1}{c}{} & \multicolumn{1}{c}{} & Person & Car & Bicycle \\
    \midrule
    MMTOD-CG\cite{MMTOD2019} & \multicolumn{1}{c}{\multirow{2}{*}{110}} & 61.40 & 63.31 & 70.63 & 50.26 \\
    MMTOD-UNIT\cite{MMTOD2019} & \multicolumn{1}{c}{} & 61.54 & 64.47 & 70.72 & 49.43 \\
    ThermalDet\cite{ThermalDet2019} & $>40.5$ & 74.60 & 78.24 & 85.52 & 60.04 \\
    ODSC\cite{ODSC2020} & $>52.6$ & 69.62 & 71.01 & 82.33 & 55.53 \\
    CFR\cite{CFR2020} & $>16.7$ & 72.39 & 74.49 & 84.91 & 55.77 \\
    GAFF\cite{GAFF2021} & 10.85 & 72.90 & - & - & - \\
    BU-ATT\cite{BU2021} & - & 73.10 & 76.10 & 87.00 & 56.10 \\
    BU-LTT\cite{BU2021} & - & 73.20 & 75.60 & 86.50 & 57.40 \\
    CMPD\cite{CMPD2022} & 110 & 69.35 & 69.64 & 78,11 & 59.87 \\
    YOLO-MSLite\cite{YOLO-MSLite} & \textcolor{green}{8.9} & 75.0 & 81.70 & 89.10 & 54.20 \\
    ICAFusion\cite{ICAFusion} & 26 & \textcolor{green}{79.20} & 81.60 & 89.00 & \textcolor{red}{66.90} \\

    \midrule
    WCCNet & 10.63 & \textcolor{red}{81.66} & \textcolor{red}{88.51} & \textcolor{red}{92.06} & \textcolor{blue}{64.42} \\
    \midrule
    WCCNet-S & \textcolor{blue}{6.74} & \textcolor{blue}{79.69} & \textcolor{blue}{87.19} & \textcolor{blue}{91.67} & \textcolor{green}{60.22} \\
    WCCNet-XS & \textcolor{red}{4.39} & 75.82 & \textcolor{green}{82.41} & \textcolor{green}{88.09} & 56.97 \\
    \bottomrule
  \end{tabular}}
  \caption{Evaluation results on FLIR dataset. \textcolor{red}{Red}, \textcolor{blue}{blue}, and \textcolor{green}{green} colors denote the first, the second, and the third place respectively. - means not reported by the paper of corresponding methods. \label{table:FLIR}}
\end{table}

\subsection{Comparison with State-of-the-arts}
\label{Exp. Comparison}

\subsubsection{Comparison on KAIST}
\label{Exp. Comparison on KAIST}
We evaluate WCCNet and conduct comparisons with other state-of-the-art methods 
\cite{MSDS2018,ARCNN2019,CIAN2019,MBNet2020,MLPD2021,CMPD2022,GAFF2021,MS-DETR2024,RPN2017Konig,zhang2023TripleNet,TCDET2020,yang2022BAANet,ICAFusion,TFNet2024}
under reasonable and all-dataset settings following \cite{KAIST2015}, as illustrated in Table \ref{Tab. KAIST} and Fig \ref{Fig. KAIST_Curve}. 
The reasonable setting means that the evaluation is conducted on a subset of testing set which only consists of pedestrians taller than 55 pixels without heavy occlusion, while all-dataset setting evaluates detectors on the whole testing set. 

\paragraph{Accuracy Comparison}
The experimental results on detection precision are given in terms of $\mathrm{MR}^{-2}$.
As can be observed in Table \ref{Tab. KAIST}, all types of our proposed WCCNet outperforms previous methods under all-dataset and reasonable settings.
Besides, the MR-FPPI curves under those two settings are shown in Fig. \ref{Fig. KAIST_Curve}(a) and Fig. \ref{Fig. KAIST_Curve}(d) respectively, which also demonstrates the superiority of WCCNet.
Especially for all-dataset setting which is more challenging, WCCNet has taken a lead of $13.71\%$ percent over the best recorded method MS-DETR \cite{MS-DETR2024} with 17.81 $\mathrm{MR}^{-2}$ for all-dataset setting.
This demonstrates the strong capability of WCCNet to handle small and occlusion pedestrians.
Moreover, for detection in low-illumination conditions, the precision results on the night subset under all-dataset setting show that WCCNet surpasses previous methods by a large margin no less than $21.56\%$. 
Therefore, it confirms that our MoWE-integrated infrared branch can extract sufficient representations of pedestrians in low-illumination conditions with remarkably fewer parameters.

\paragraph{Efficiency Comparison}
For evaluation on model efficiency, the number of parameters, FLOPs, and the inference time are provided in Table. \ref{Tab. KAIST}.
Besides, the Speed vs. $\mathrm{MR}^{-2}$ curves under reasonable setting are shown in Fig. \ref{fig. Speed vs. Accuracy}.
It is obvious that, compared with previous works, WCCNet achieves remarkable fast speed with fewer parameters and lower computational cost while maintaining superior accuracy. 
Especially, WCCNet-S with only 9.75$M$ parameters and 53.86$G$ FLOPs can infer at a speed of $152.44$ frames per second (FPS) while maintain competitive accuracy, which again shows the efficiency of WCCNet and its potential in practical applications.

\begin{figure}[!t]
    \centering
    \includegraphics[width=0.9\linewidth]{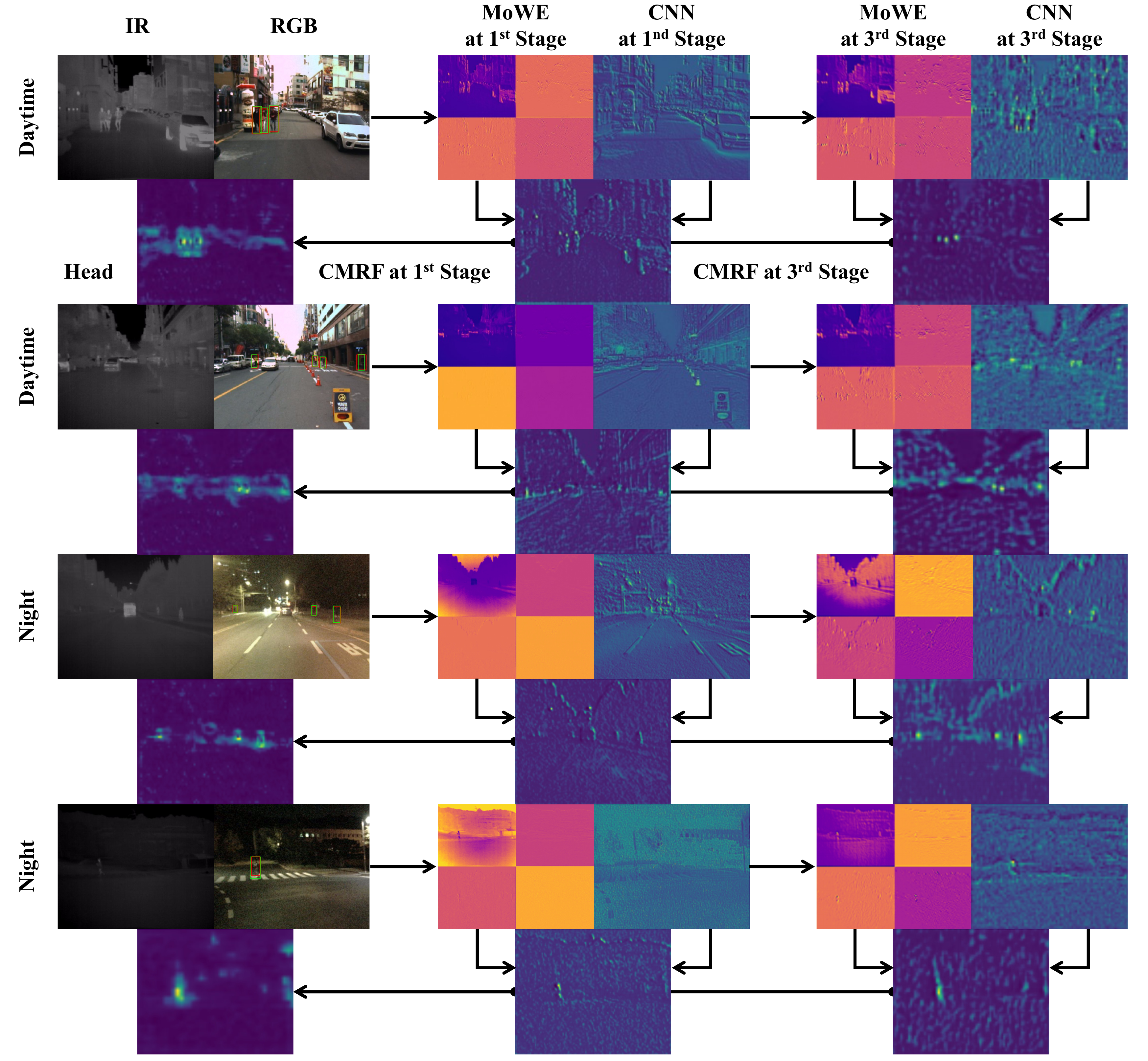}
    \caption{Visualization of extracted and integrated multispectral features of WCCNet. Origin images, features extracted by different branches, fused features by CMRF, and the head features for prediction are shown above, which are grouped by daytime and night. Limited by space, only features at $1^{st}$ and $3^{rd}$ stage are visualized here. As can be observed, pedestrian features are effectively extracted by two different branches, and are further aggregated and refined by CMRF to yield discriminative head features for accurate prediction.\label{Fig. Features in Backbones}}
\end{figure}

\paragraph{Visual Comparison}
Several detection results for complex scenarios are shown in Fig. \ref{Fig. det_result}.
It can be observed that WCCNet is able to accurately detect pedestrians in challenging situations, including heavy occlusions, absence of RGB information with low illumination, and small-scale pedestrians.

\subsubsection{Comparison on FLIR}

In order to illustrate the versatility of WCCNet when handling other categories, further evaluation is conducted on FLIR which contains multiple types of objects.
The experimental results are shown in Table \ref{table:FLIR}.
Precision results are measured using AP50 for each class and mAP50 over all classes with an IoU threshold of 0.5, which remains the same as in previous works.
Table \ref{table:FLIR} shows that WCCNet gets the best result with $81.66\%$ mAP for overall mAP50, outperforming previous works by at least $9.46\%$.
The results demonstrate that our WCCNet is able to generalize well to objects of different categories, not only pedestrians.

\subsection{Ablation Study \& Further Analysis\label{Exp.ablation}}
In this subsection, we perform ablation experiments on KAIST dataset for detailed analysis of WCCNet.
The training settings and other unmentioned hyperparameters are the same across all variants of WCCNet for fair comparison.

\subsubsection{Analysis of Wavelet-Cooperative Backbone\label{Exp.dual_backbone}}

\paragraph{Different Backbone Types}
To confirm the efficiency of the proposed dual-stream backbone, which extracts infrared features by MoWE and RGB features by general neural layers, the evaluation on the performance of different backbone combinations has been carried out and summarized in Table \ref{Tab. branch}.
The backbone with dual CNN subnetworks in the first row of Table \ref{Tab. branch} simply replaces the origin MoWE subnetwork in infrared branch with CNN subnetwork.
For the backbone with dual MoWE subnetworks in the second row of Table \ref{Tab. branch}, two symmetrical CE layers are applied for both subnetworks in CMRF module.
The backbone of WCCNet is shown in the gray row.

\begin{table}[!t]
  \centering
  \resizebox{\textwidth}{!}{
  \renewcommand{\arraystretch}{1}
  \begin{tabular}{cc|ccc|ccc|ccc|ccc}
    \hline\hline
    \multicolumn{2}{c|}{Input Modalities}&\multicolumn{3}{c|}{Backbone Types for Branches}& \multicolumn{3}{c|}{Time Complexity \& Space Complexity}& \multicolumn{3}{c|}{$\mathrm{MR}^{-2}$ (All)}& \multicolumn{3}{l}{$\mathrm{MR}^{-2}$ (Reasonable)}\\
    \cline{1-2}\cline{3-5}\cline{6-8}\cline{9-11}\cline{12-14}
    RGB & Infrared & RGB Branch & Infrared Branch & Fusion & Params. (M)& FLOPs (G)&Time (ms)& All& Day& Night& All& Day& night\\

    \cline{1-14}
    \Checkmark & \Checkmark  & CNN & CNN  & CMRF & 25.23  & 130.82  & 16.91   & 19.02  & 21.77  & 13.26  & 7.96   & 9.54   & 5.09  \\

    \Checkmark & \Checkmark  & MoWE & MoWE  & CMRF & 10.00  & 48.96  & 5.13     & 24.44  & 28.90  & 16.31  & 10.59   & 12.49   & 7.11  \\

    \Checkmark & \Checkmark  & MoWE & CNN & CMRF & 17.50  & 91.00  & 10.62     & 20.07  & 24.09  & 12.72  & 8.55   & 10.54   & 4.92  \\
    
    \rowcolor{gray!30}
    \Checkmark & \Checkmark  &CNN & MoWE & CMRF & 17.50  & 91.00  & 10.62   & 17.81  & 20.36  & 12.37  & 6.72   & 7.80   & 4.72 \\
    
    \cline{1-14}
    \Checkmark & \XSolidBrush  &CNN & \XSolidBrush & \XSolidBrush & 16.71  & 64.93  & 5.70   & 42.90  & 35.53  & 60.58  & 22.62   & 15.69   & 36.61  \\

    \XSolidBrush & \Checkmark  &\XSolidBrush & MoWE & \XSolidBrush & 9.77  & 46.03  & 4.42   & 29.80  & 36.45  & 17.65  & 14.36   & 17.69   & 8.28  \\
    \hline\hline
  \end{tabular}}
    \caption{Accuracy and efficiency evaluation of different backbone combinations applied in RGB/Infrared branch. The conditions with block-out input modality are also evaluated, where \XSolidBrush \; means the block-out modules or modalities.\label{Tab. branch}}
\end{table}

Table \ref{Tab. branch} reveals an apparent $12.69$\% decrease in precision when reversing the paired CNN-MoWE subnetworks of WCCNet.
Moreover, there are a noteworthy $37.20$\% reduction in inference time and a $6.36$\% increase in accuracy when backbone with dual CNN subnetworks is replaced with backbone with paired CNN-MoWE subnetworks. 
Therefore, MoWE is more efficient to extract infrared features than CNN.
It is also evidenced in Fig. \ref{Fig. Features in Backbones} that, compared with the infrared features extracted by CNN, the approximation features extracted by MoWE exhibit more distinct and clearer responses to targets both in daytime and at night.
However, a significant decline in accuracy occurs when dual-MoWE subnetworks are applied, which is further supported by daytime scenarios in Fig. \ref{Fig. Features in Backbones} that CNN subnetwork extracted more distinguished features for RGB modality than MoWE.
The above observation shows the great significance to asymmetrically build dual-stream wavelet-cooperative backbone which differentially extracts features in different domain.

\begin{table}[!t]
  \centering
\resizebox{\textwidth}{!}{
  \renewcommand{\arraystretch}{1}
  \begin{tabular}{ccc|ccc|cc}
    \hline\hline
    \multicolumn{3}{c|}{MoWE Configurations}&\multicolumn{3}{c|}{$\mathrm{MR}^{-2}$ (All Setting)}& \multicolumn{2}{c}{Computational Cost}\\
    \cline{1-3}\cline{4-6}\cline{7-8}
    Routed Expert Number $N_e$ & Balance Loss $\mathcal{L}_{b}$ & Shared Expert & All & Day & Night & Params. (M)& FLOPs (G)\\
    \cline{1-8}
    0 & \Checkmark & \Checkmark & 19.21 & 21.98 & 13.42 & 17.490458 & 89.844787 \\
    2 & \Checkmark & \Checkmark & 18.65 & 21.77 & 12.95 & 17.492704 (+0.002246) & 90.169361 (+0.324574) \\
    4 & \Checkmark & \Checkmark & 18.12 & 21.14 & 12.60 & 17.493550 (+0.003092) & 90.447597 (+0.602810) \\
    \hline
    8 & \Checkmark & \XSolidBrush & 18.44 & 21.32 & 13.18 & 17.495210 (+0.004752) & 90.946725 (+1.101938) \\
    8 & \XSolidBrush & \Checkmark & 18.63 & 21.55 & 13.30 & 17.495242 (+0.004784) & 91.004069 (+1.159282) \\
    \rowcolor{gray!30}
    8 & \Checkmark & \Checkmark & 17.81 & 20.36 & 12.37 & 17.495242 (+0.004784) & 91.004069 (+1.159282) \\
    \hline
    16 & \Checkmark & \Checkmark & 17.55 & 20.65 & 11.89 & 17.498650 (+0.008192) & 92.117013 (+2.272226) \\
    32 & \Checkmark & \Checkmark & 17.43 & 20.32 & 10.73 & 17.505394 (+0.014936) & 94.342903 (+4.498116) \\
    \hline\hline
  \end{tabular}}
    \caption{Accuracy and efficiency evaluation with different expert configurations. Metrics are evaluated on KAIST test set. The activation ratio of routed experts is set to 0.5.\label{table:expert_config}}
\end{table}

\paragraph{Visualization of Feature Extraction and Cooperation} We visualize extracted features of different branches, fused features by CMRF, and the head features for prediction in Fig. \ref{Fig. Features in Backbones}.
It can be observed that, for night scenarios, pedestrian features extracted by MoWE are more prominent than those extracted by CNN, which CNN achieve better at daytime.
Besides, compared with solely extracted features in one branch, features fused by CMRF exhibit more concentrated responses to pedestrians, which demonstrates the effectiveness of CMRF.
The final head features show that WCCNet can accurately locate pedestrians for complex scenarios.

\begin{figure}[!t]
    \centering
    \includegraphics[width=0.75\linewidth]{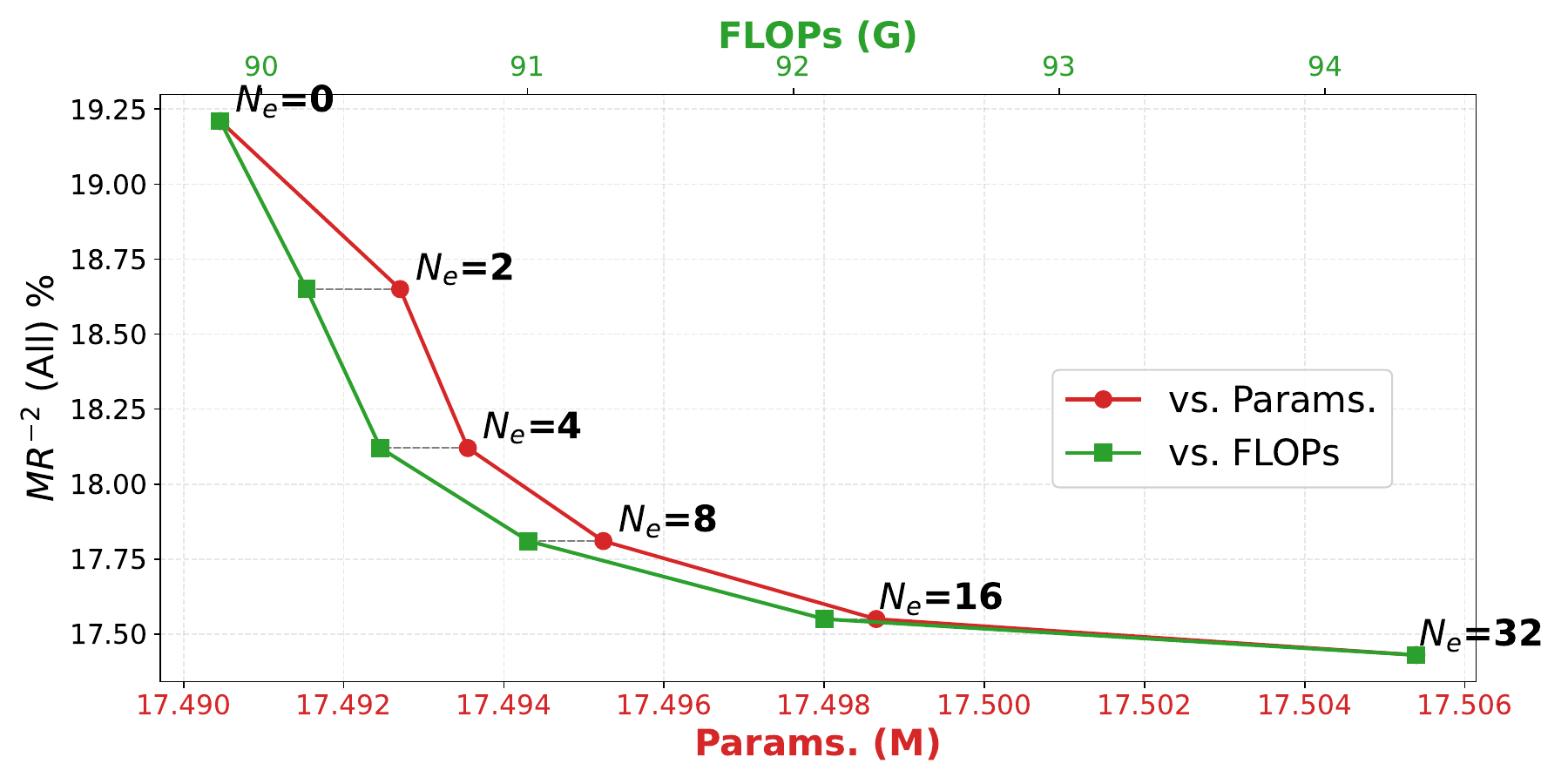}
    \caption{Accuracy vs. efficiency analysis of varying routed expert numbers. Detection Accuracy $MR^{-2}$ is compared with model parameters and FLOPs.
    \label{fig:moe_number}}
\end{figure}

\subsubsection{Analysis of MoWE\label{sec: MoWE Exp}}

\paragraph{Analysis of Different Expert Configurations}
We conduct ablation experiments on three main configurations of MoWE, i.e., routed expert number $N_e$, balance loss $\mathcal{L}_{b}$, and shared expert, as shown in Table \ref{table:expert_config} and Fig. \ref{fig:moe_number}.
There is an improvement of $7.29\%$ in accuracy when $N_e$ change from 0 to 8, demonstrating the significance of routed expert ADWT.
As the number of routed experts $N_e$ increases from 0 to 32, the accuracy and computational cost simultaneously increase.
Fig. \ref{fig:moe_number} shows that $N_e$=8 meets a trade-off between accuracy and efficiency.
Besides, compared with generic neural layers, the increment of computational overhead brought by MoWE is marginal, i.e., less than 0.08\% in paramters and 5.01\% in FLOPs, which highlights the efficiency of MoWE.

The involvement of balance loss $\mathcal{L}_{b}$ and shared expert can further improve the accuracy by $3.42\%$ and $4.40\%$ respectively.
The balance loss help to prevent routing collapse and fully explore all the available experts, while the shared expert can further enhance the feature extraction stability of MoWE.

\begin{figure}[!t]
    \centering
    % --- 第一行：图 A ---
    \subfloat[]{
        \centering
        \includegraphics[width=0.92\linewidth]{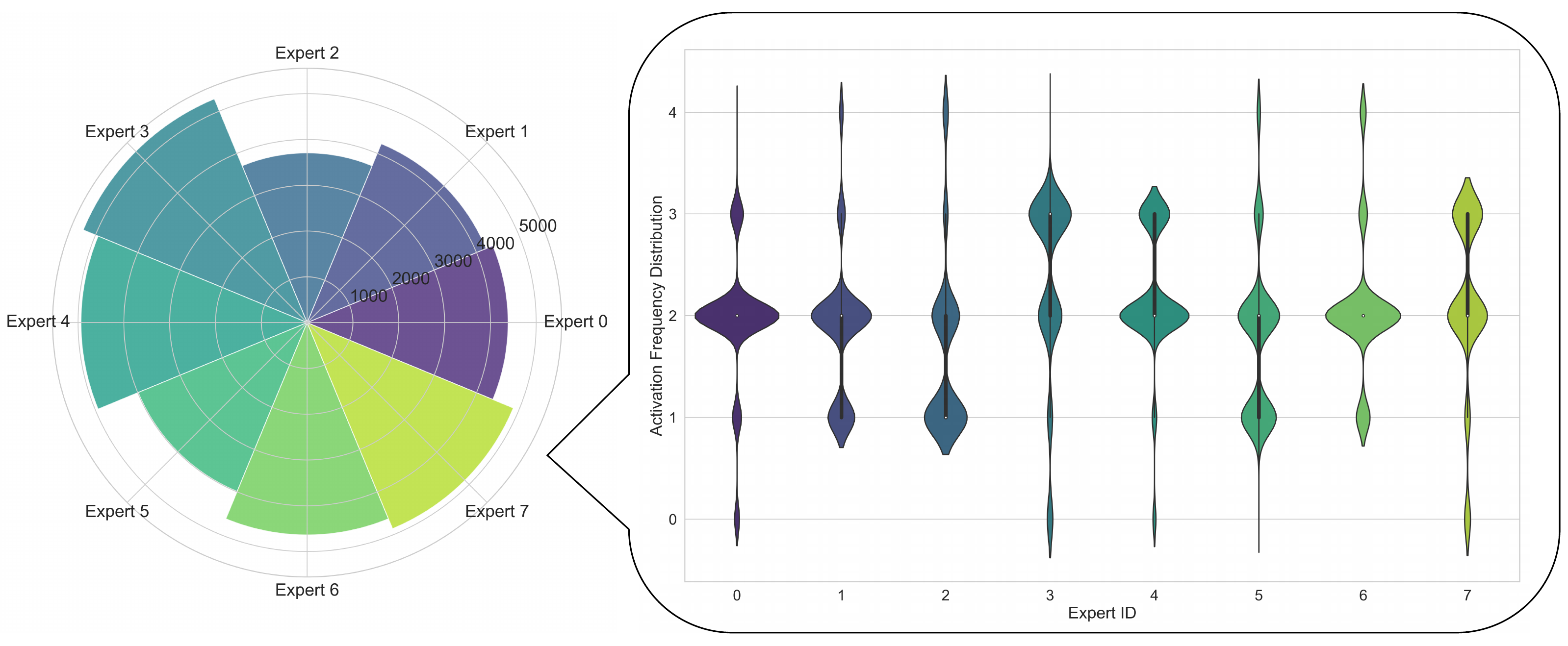}
        \label{fig:moe_sub_a}
    }

    % --- 第二行：图 B 和 图 C ---
     \subfloat[]{
        \centering
        \includegraphics[width=0.36\linewidth]{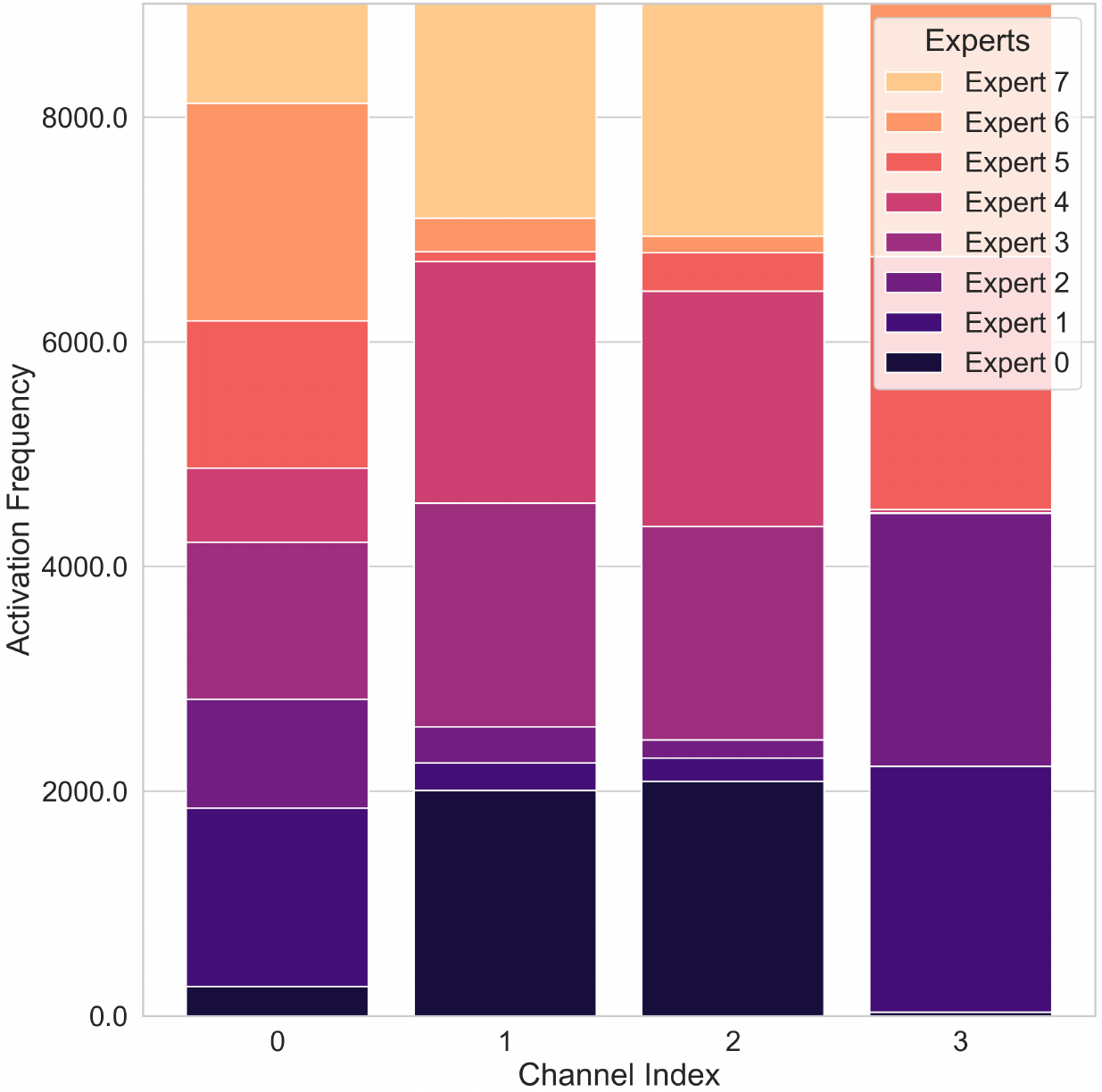}
        \label{fig:moe_sub_b}
    }
    \subfloat[]{
        \centering
        \includegraphics[width=0.56\linewidth]{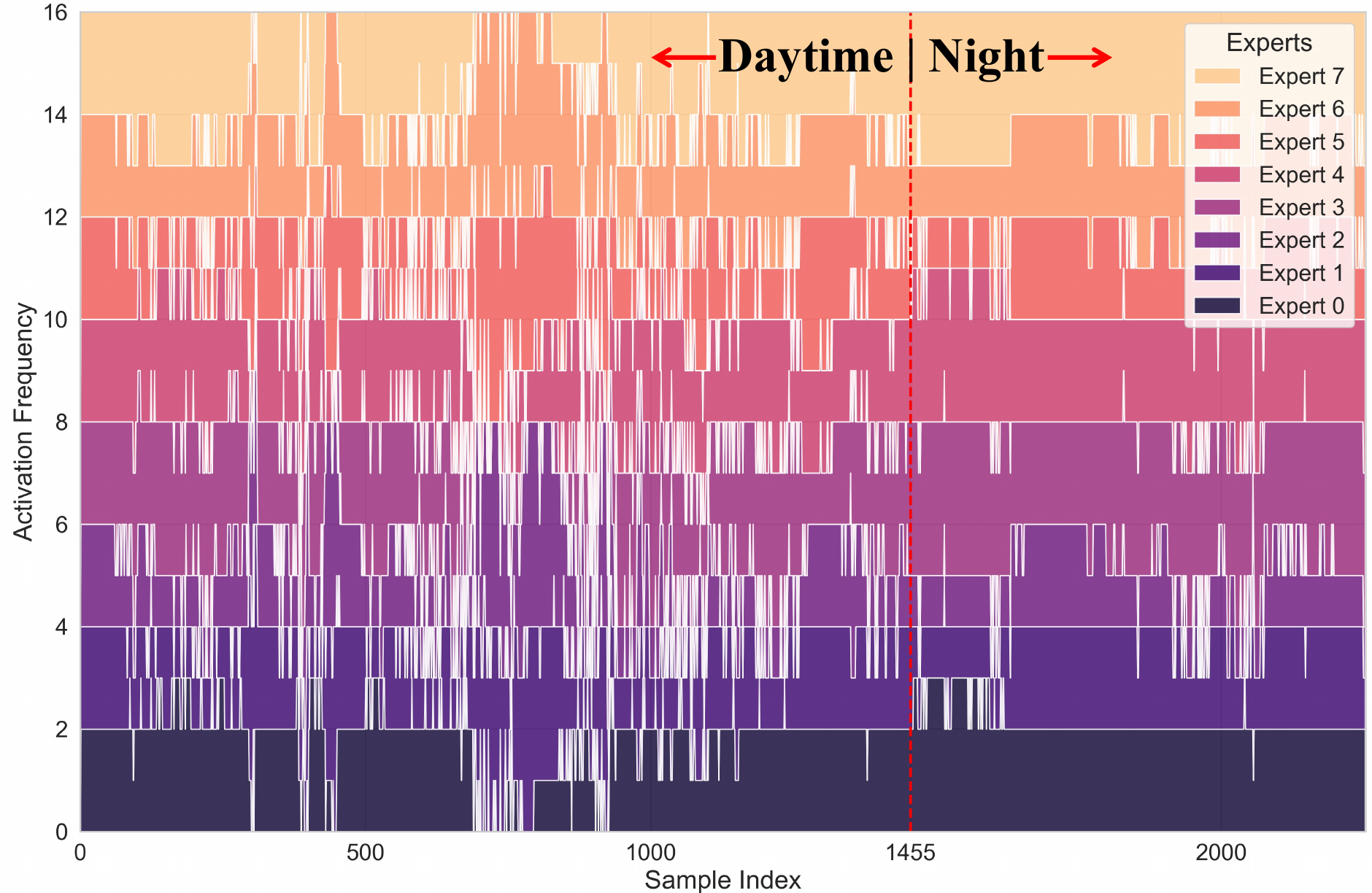}
        \label{fig:moe_sub_c}
    }
    
    \caption{Analysis of routed expert activation patterns for MoWE over the entire KAIST test set. (a) compares the activated ratio and corresponding distribution across different experts in left and right sub-figures respectively. (b) illustrates the activation frequency distribution across different channels. (c) shows the activation frequencies of experts across samples for daytime and night scenarios.\label{fig:moe_analysis}}
\end{figure}

\paragraph{Analysis of Expert Activation Patterns}
Fig. \ref{fig:moe_sub_a}, Fig. \ref{fig:moe_sub_b}, and Fig. \ref{fig:moe_sub_c} show the activation patterns across different experts, channels, and samples respectively.
Fig. \ref{fig:moe_sub_a} shows that the activation ratio of each expert is approximately equal, which demonstrates that the routing mechanism is balanced.
Fig. \ref{fig:moe_sub_b} indicates that each channel has their preferred experts.
That is, different experts have their own strengths in feature extraction for different channels.
It can be observed from Fig. \ref{fig:moe_sub_c} that the activation patterns of experts vary distinctively between daytime and night-time samples.
The activation frequencies become more stable across night-time samples, and Expert $\{3,4,7\}$ exhibit higher usage.
This behavior is attributed to the fact that night-time infrared images possess more salient pedestrian features, which are easier for MoWE to capture than their daytime counterparts.

\begin{figure}[!t]
  \centering
  \includegraphics[width=\textwidth]{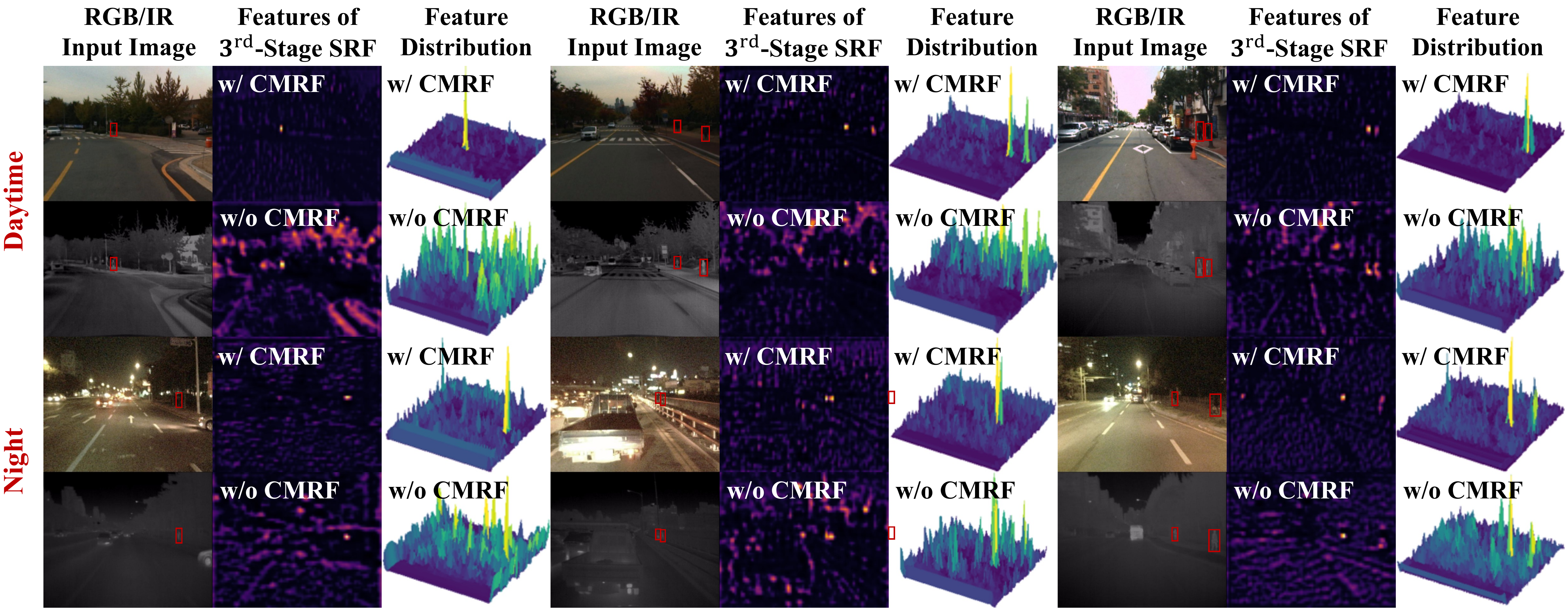}
  \caption{Comparison on fused features $\mathbf{G}_{T}^j$ with/without CMRF. Fused $\mathbf{G}_{T}^j$ features in CMRF are visualized by heatmaps and 3D curves in different columns, which are grouped by different times including daytime and night. When not applying CMRF, $\mathbf{G}_{T}^j$ refers to features of Mid-Cat fusion. As can be observed, feature maps of CMRF effectively concentrate on spatial regions associated with pedestrians, suppressing interference from similar backgrounds. \label{Fig. vis_CMRF}}
\end{figure} 

\begin{table}[!h]
  \centering
%   \large
  \renewcommand{\arraystretch}{1}
  \resizebox{0.7\textwidth}{!}{
  \setlength{\tabcolsep}{3pt}
  \begin{tabular}{cccccc}
  \hline\hline
  \multirow{2}{*}{} & \multicolumn{5}{c}{The Stages that CNN Interchanged with ADWT} \\
  \cline{2-6}
   & Stage 1 & Stages 1-2 & Stages 1-3 & Stages 1-4 & Stages 1-5 \\
  \hline
  \multicolumn{1}{c}{\cellcolor[HTML]{D8D8D8}Datasets} & \multicolumn{5}{c}{\cellcolor[HTML]{D8D8D8}Precision} \\
  \hline
  MNIST & 0.9964 & 0.9964 & 0.995 & 0.9936 & 0.9875 \\
  FashionMNIST & 0.9382 & 0.9368 & 0.9329 & 0.9172 & 0.8862 \\
  CIFAR-10 & 0.9137 & 0.9142 & 0.8885 & 0.8173 & 0.6206 \\
  \hline
  \multicolumn{1}{c}{\cellcolor[HTML]{D8D8D8}Metrics} & \multicolumn{5}{c}{\cellcolor[HTML]{D8D8D8}Efficiency} \\
  \hline
  FLOPs (G) & 1.920 & 1.600 & 0.993 & 0.544 & 0.194 \\
  Params. (M) & 22.2 & 22.0 & 20.4 & 16.0 & 1.96\\
  \hline\hline
  \end{tabular}}
\caption{Evaluation on ADWT-integrated backbone variants which interchange CNN layers with ADWT layers at different stages. Grayscale and RGB images of three classification benchmarks are involved.\label{table:ADWT ClS}}
\end{table}

\paragraph{Effectiveness of ADWT-integrated Neural Network}
To evaluate the efficiency of ADWT, we extract RGB branch of WCCNet as classification model, and progressively replaced CNN layers with ADWT layers from stages 1-5. 
Evaluations are conducted on grayscale datasets MNIST and FashionMNIST, and RGB dataset CIFAR-10, as shown in Table \ref{table:ADWT ClS}.
Increasing ADWT substitution significantly reduces complexity, up to $89.9\%$ FLOPs and $91.2\%$ parameters, but trades off precision. 
Grayscale datasets show minimal accuracy degradation, $0.89\%$ for MNIST and $5.54\%$ for FashionMNIST, benefiting from frequency-domain processing of ADWT. 
However, RGB data (CIFAR-10) suffered a $32.08\%$ drop in deeper substitutions despite a tenfold efficiency gain, indicating ADWT is more effective for grayscale than RGB feature extraction.

\subsubsection{Analysis of Crossmodal Rearranging Fusion\label{sec: Exp.CMRF}}

\paragraph{Quantitative Comparisons with Different Fusion Schemes}
Table \ref{Tab. Fusion} shows ablation studies on comparison between our proposed CMRF and other various fusion schemes. 
The different fusion schemes are all implemented within WCCNet framework.
The Early Concatenation (Early-Cat) and Late Concatenation (Late-Cat) fusion schemes only concatenate multimodality feature maps at the first and the last stage of the backbone along the channel dimension respectively, and get the lowest precision.
The Middle Concatenation (Mid-Cat) fusion scheme concatenates RGB and infrared features along the channel dimension at every stage of the backbone, and thus obtain higher accuracy with a minor decline in $MR^{-2}$.
These channel-dimension concatenation-based fusion schemes neglect the interaction between multimodal features which is essential for extracting complementary information and filtering out redundant information, hence unsurprisingly they get the lowest precision.

The middle fusion schemes designed for crossmodal interaction including DMAF, Inter-MA, and CMRF, are also evaluated.
It is worth mentioning that the features for middle fusion are all extracted after CE layer.
The DMAF\cite{MBNet2020} and Inter-MA\cite{zhang2023TripleNet} both obtain complementary information by subtracting feature maps of different modalities along the channel dimension to extract numerically differential features.
Instead, the proposed CMRF extracts complementary information through mining semantic-aware features in spatially relevant local regions shared by different modalities.

Among the above fusion schemes, CMRF achieves the highest accuracy with competitive inference time, outperforming the second-best DMAF by $13.71\%$ in accuracy.

\begin{table}[!t]
  \centering
  \renewcommand{\arraystretch}{1}
  
  \resizebox{0.7\textwidth}{!}{
  \begin{tabular}{ccc|cc|ccc}
  \hline
  \hline
  \multicolumn{3}{c|}{\multirow{2}{*}{\tabincell{c}{Fusion Scheme for WCCNet}}} & \multicolumn{2}{c|}{Efficiency} & \multicolumn{3}{c}{$\mathrm{MR}^{-2}$ (All)} \\ \cline{4-8} 
  \multicolumn{3}{c|}{} & Params. (M) & Time (ms) & All & Day & Night \\ \hline
  \multicolumn{8}{c}{\cellcolor[HTML]{D8D8D8}Early Fusion}\\
  \hline
  \multicolumn{3}{c|}{Early-Cat\cite{RPN2017Konig}} &  16.71& 6.91 & 20.99 & 23.27 & 15.04 \\
  \hline
  \multicolumn{8}{c}{\cellcolor[HTML]{D8D8D8}Middle Fusion}\\
  \hline
  \multicolumn{3}{c|}{Mid-Cat\cite{MLPD2021}} &16.76  & 6.14  & 20.93 & 23.54 & 14.04 \\
  \multicolumn{3}{c|}{DMAF\cite{MBNet2020}} &17.51  & 8.38 & 20.64 & 23.29 & 14.75 \\
  \multicolumn{3}{c|}{Inter-MA\cite{zhang2023TripleNet}} & 17.13 &7.24  & 21.54 & 24.80 & 14.20 \\
  \hline
  \multicolumn{8}{c}{\cellcolor[HTML]{D8D8D8}Late Fusion} \\
  \hline
  \multicolumn{3}{c|}{Late-Cat\cite{RPN2017Konig}} &16.80  & 6.02 & 22.67 &25.04  &17.98  \\
  \hline
  \multicolumn{8}{c}{\cellcolor[HTML]{D8D8D8}Proposed CMRF}\\ 
  \cline{1-8}
  SRF & CE & CSA & Params. (M) & Time (ms) & All & Day & Night \\ \hline
 \Checkmark& \XSolidBrush & \XSolidBrush & 17.16  & 8.82  & 19.35 & 22.50 & 13.60 \\
  \Checkmark & \Checkmark & \XSolidBrush & 17.18  & 8.91  & 18.25 & 21.73 & 11.90 \\
  \Checkmark &  \XSolidBrush & \Checkmark & 17.49  & 10.61  & 18.85 & 22.05 & 13.01 \\
  \Checkmark & \Checkmark & \Checkmark & 17.50  & 10.62  & 17.81 & 20.36 & 12.37\\
  \hline
  \hline
  \end{tabular}
  }
    \caption{Evaluation of different fusion settings for WCCNet, along with ablation study on the proposed components of CMRF. Accuracy is evaluated by $\mathrm{MR}^{-2}$ with All setting.\label{Tab. Fusion}}
 \end{table}

\paragraph{Analysis of Different Components in CMRF}
The influence of different components of CMRF is investigated in Table \ref{Tab. Fusion}.
The baseline is set to that only Semantic Rearranging Fusion(SRF) is applied in WCCNet, while achieving $20.20$ $MR^{-2}$.
The accuracy is improved to $19.53$ $MR^{-2}$ when Cross-domain Embedding(CE) layers are added to narrow the semantic gap of multimodal features before being fused by SRF. 
Lastly, the Cross-domain Spatial Alignment(CSA) is adopted, and then the accuracy further reaches $19.05$ $MR^{-2}$ as features of different spectra are aligned.
Each component of CMRF indeed contributes to the final accuracy.

\paragraph{Visual Illustration for CMRF}
As shown in Fig. \ref{Fig. vis_CMRF}, for difficult scenarios that pedestrians not clearly visible in any single modality or unrelated target interference, CMRF effectively
focuses on the regions relevant to pedestrians.
This observation demonstrates that the complementary features of pedestrians in spatially-related local regions for different modalities are effectively aggregated by CMRF.

\balance

\section{Conclusion}
In this paper, we have proposed a novel framework WCCNet to achieve efficient multispectral pedestrian detection. 
Based on wavelet-cooperative backbone integrated with efficient MoWE layers, WCCNet has boosted the inference speed by learning spectra-specific representations while maintaining competitive accuracy.
Based on environment context, MoWE leverages shared static DWT and  selectively activated ADWT layers to efficiently extract rich and informative features from infrared images with low computational overhead.
For further exploration on multispectral reciprocal cooperation, CMRF has been proposed to merge the spatially relevant local regions shared by different spectra with adaptive weights learned on-the-fly. 
Extensive experiments on KAIST and FLIR datasets have been conducted and substantiate the aforementioned superiority of WCCNet over several state-of-the-art methods.
We believe that WCCNet can potentially benefit the development of resource-constrained detection systems for autonomous driving applications.
In the future, we aim to further enhance the representation capability of MoWE and the fusion efficiency of CMRF, while extending the WCCNet framework to diverse backbones and broader multimodal tasks.
% Our ablation studies also revealed certain limitation of WCCNet.
% While MoWE proves highly efficient for infrared images, it is less effective for texture-rich RGB data, necessitating the involvement of heavier generic neural layers. 
% Further reduction of computational overhead can be explored in the future for the improvement of MoWE.
% In general, WCCNet offers a promising solution for developing resource-constrained perception systems in autonomous driving field.


%% Use figure environment to create figures
%% Refer following link for more details.
%% https://en.wikibooks.org/wiki/LaTeX/Floats,_Figures_and_Captions
% \begin{figure}[t]%% placement specifier
% %% Use \includegraphics command to insert graphic files. Place graphics files in 
% %% working directory.
% \centering%% For centre alignment of image.
% \includegraphics{example-image-a}
% %% Use \caption command for figure caption and label.
% \caption{Figure Caption}\label{fig1}
% %% https://en.wikibooks.org/wiki/LaTeX/Importing_Graphics#Importing_external_graphics
% \end{figure}

% \appendix
% \section{Example Appendix Section}


%%       \cite{<label>} ==> [1]

%%

% \clearpage
\bibliographystyle{elsarticle-num} 
\bibliography{references.bib}

\end{document}